\documentclass{article} % For LaTeX2e
\usepackage{iclr2026_conference,times}

\usepackage{amsmath,amsfonts,bm}

\def\eqref#1{equation~\ref{#1}}
\def\1{\bm{1}}

\DeclareMathAlphabet{\mathsfit}{\encodingdefault}{\sfdefault}{m}{sl}
\SetMathAlphabet{\mathsfit}{bold}{\encodingdefault}{\sfdefault}{bx}{n}

\usepackage{microtype}
\usepackage{graphicx}
\usepackage{xcolor}
\usepackage{subcaption}
\usepackage{booktabs} % for professional tables
\usepackage{tabularx} % Add this to your preamble
\usepackage{hyperref}
\usepackage{multirow}
\usepackage{amsmath}
\usepackage{amssymb}
\usepackage{mathtools}
\usepackage{amsthm}

\usepackage{algorithm}
\usepackage{algpseudocode}
\usepackage{bm} % Added to fix the \bm undefined control sequence error

\usepackage[capitalize,noabbrev]{cleveref}

\theoremstyle{plain}

\theoremstyle{definition}

\theoremstyle{remark}

\usepackage[textsize=tiny]{todonotes}

\usepackage{colortbl}
\definecolor{best}{RGB}{185,200,230}      % 1st place, darker blue
\definecolor{runnerup}{RGB}{225,235,250}  % 2nd place, lighter blue

\newcommand{\first}[1]{\cellcolor{best}\textbf{#1}}
\newcommand{\second}[1]{\cellcolor{runnerup}#1}

\title{ROSER: Few-Shot Robotic Sequence Retrieval for Scalable Robot Learning}

\author{Zillur Rahman
\thanks{Correspondance to: zillur.mle@gmail.com}\\
University of Nevada, Las Vegas\\
\And
Eddison Pham\\
University of Toronto\\
\And
Alejandro Daniel Noel\\
LatentWorlds AI, NL\\
\AND
Cristian Meo
\thanks{Advisor: cristianmeo@latentworlds.ai, Website: \url{www.latentworlds.ai}}\\
LatentWorlds AI, NL\\
}

\iclrfinalcopy % Uncomment for camera-ready version, but NOT for submission.
\begin{document}

\maketitle

\begin{abstract}
A critical bottleneck in robot learning is the scarcity of task-labeled, segmented training data, despite the abundance of large-scale robotic datasets recorded as long, continuous interaction logs. Existing datasets contain vast amounts of diverse behaviors, yet remain structurally incompatible with modern learning frameworks that require cleanly segmented, task-specific trajectories. We address this data utilization crisis by formalizing robotic sequence retrieval: the task of extracting reusable, task-centric segments from unlabeled logs using only a few reference examples. We introduce ROSER, a lightweight few-shot retrieval framework that learns task-agnostic metric spaces over temporal windows, enabling accurate retrieval with as few as 3-5 demonstrations, without any task-specific training required. To validate our approach, we establish comprehensive evaluation protocols and benchmark ROSER against classical alignment methods, learned embeddings, and language model baselines across three large-scale datasets (e.g., LIBERO, DROID, and nuScenes). Our experiments demonstrate that ROSER consistently outperforms all prior methods in both accuracy and efficiency, achieving sub-millisecond per-match inference while maintaining superior distributional alignment. By reframing data curation as few-shot retrieval, ROSER provides a practical pathway to unlock underutilized robotic datasets, fundamentally improving data availability for robot learning. %\footnote[1]{Correspondance to: zillur.mle@gmail.com, cristianmeo@latentworlds.ai, Website: \url{www.latentworlds.ai}}
\end{abstract}

\section{Introduction}

The promise of generalist robot learning depends critically on access to massive, diverse datasets spanning tasks, embodiments, and environments \citep{brohan2023rtx}. Yet despite significant progress in large-scale data collection \citep{caesar2020nuscenes,liu2023libero,khazatsky2024droid}, a fundamental bottleneck persists: the vast majority of existing robotic data remains structurally incompatible with modern learning frameworks (e.g., Vision-Language-Action models \citep{shao2025vlabotsurvey}, World Models \citep{meo2024masked}). While state-of-the-art approaches demand cleanly segmented, task-labeled trajectories \citep{mandlekar2019scaling, Ebert-RSS-22}, real-world robot logs are recorded as long, continuous streams lacking task boundaries, semantic labels, or hierarchical annotations \citep{mandlekar2019scaling,Ebert-RSS-22}.

This structural mismatch creates a data utilization crisis. Datasets like nuScenes \citep{caesar2020nuscenes}, DROID \citep{khazatsky2024droid}, and LIBERO \citep{liu2023libero} contain thousands of hours of diverse behaviors, yet extracting reusable task segments requires prohibitively expensive human annotation or domain-specific heuristics that do not generalize \citep{STRAP2025,Kumar2025COLLAGE}.

We propose re-framing data curation as a few-shot retrieval problem. Rather than requiring dense supervision, a small number of reference demonstrations should suffice to identify all semantically similar segments within large, unlabeled datasets. This paradigm shift eliminates exhaustive annotation, enables rapid adaptation to new tasks, and naturally handles execution variability through learned similarity rather than brittle template matching.

However, robotic sequence representation poses fundamental challenges. Unlike images or text, robot trajectories are variable-length, high-dimensional time series with complex temporal dependencies, control noise, and execution stochasticity \citep{Argall2009SurveyLfD,ArcLengthWarping2024}. Existing approaches fall short: classical trajectory alignment methods like dynamic time warping (DTW) lack semantic understanding \citep{sako78dynamic}, embedding-based approaches remain sensitive to execution variability \citep{TimewarpVAE2023,STRAP2025,STTraj2Vec2024}, and language or vision-guided methods \citep{zeng2023llmrobotics,shao2025vlabotsurvey} struggle with fine-grained kinematic structure. Recent work on trajectory representations emphasizes learning compact embeddings that capture spatio-temporal and task-level semantics \citep{STTraj2Vec2024,STRAP2025}, yet these methods typically require task-specific fine-tuning or dense supervision.

Drawing on metric-based few-shot learning \citep{snell2017prototypical,koch2015siamese,sung2018learning}, we introduce Robotic Sequence Retrieval (ROSER), a lightweight framework that learns task-agnostic metric spaces directly from raw proprioceptive time series. ROSER constructs prototype representations from a handful of reference examples and retrieves similar segments via learned distances over short temporal windows, requiring no task-specific training at deployment. This approach is based on Prototypical Networks \citep{snell2017prototypical}, which represent each task by means of the embedding of its support examples, enabling closest-prototype comparisons that naturally support the quick retrieval of new tasks. To the best of our knowledge, this is among the first works to study few-shot robotic sequence retrieval using only proprioceptive time-series signals.

Our main contributions are: (C1) Formalizing robotic sequence retrieval and proposing ROSER, a few-shot framework that enables accurate retrieval with 3-5 examples.(C2) Establish evaluation protocols and benchmarking across three large-scale datasets against classical, learned, and foundation model baselines. (C3) Demonstrating that ROSER outperforms all baselines in accuracy and efficiency, providing a practical solution to unlock underutilized robotic data at scale.

\section{Related Works}
\textbf{Data Retrieval in Robotics and Autonomous Driving.} Large-scale robotic and driving datasets enable advances in manipulation, navigation, and scene understanding, yet retrieving task-relevant segments remains challenging. Methods such as STRAP \citep{STRAP2025} and COLLAGE \citep{Kumar2025COLLAGE} rely on heuristics, which are often dataset-specific and not scalable. Visual place recognition (VPR) surveys \citep{Lowry2015VPRsurvey, Masone2021DeepVPRsurvey} and temporal contrastive learning \citep{Dave2021TCLR} show that learned embeddings, static or temporal, can improve the retrieval of semantically meaningful segments. Multi-modal embeddings like CLIP \citep{Radford2021CLIP} further enable few-shot, cross-domain generalization. In autonomous driving, BEV-based retrieval \citep{Tang2024BEVTSRTRA} requires multi-view images and large labeled datasets, while front-facing models for action or scene classification \citep{Noguchi2023EgoVehicleAR, vellenga2025latentuncertaintyrepresentationsvideobased, Vellenga2024EvaluationOV, Wang2025DrivingBR} are data- and compute-intensive. In contrast, we aim to retrieve scenes using only a few labeled samples with a lightweight, generalizable model suitable for resource-constrained environments.

\textbf{Few-Shot Learning.}  
Few-Shot Learning (FSL) addresses the gap between human-like learning, which requires few examples, and traditional deep learning, which relies on massive annotated datasets. Meta-learning is a prominent paradigm in FSL. \citet{koch2015siamese} introduced Siamese Networks to learn similarity functions between image pairs. \citet{snell2017prototypical} proposed Prototypical Networks, representing each class by the mean of its examples in embedding space, enabling nearest-prototype classification. \citet{finn2017} developed Model-Agnostic Meta-Learning (MAML), which finds parameter initializations optimized for fast adaptation to new tasks. More recently, CLIP \citet{radford2021learningtransferablevisualmodels} demonstrated that large-scale image-text pretraining enables strong zero- and few-shot transfer by aligning visual and linguistic embeddings. These approaches provide the foundation for few-shot retrieval of driving scenes from minimal labeled data.

\textbf{Time-Series Motifs.}  
Time-series motifs are recurring, characteristic subsequences within sequential data that can capture meaningful patterns without requiring full supervision. Early approaches such as shapelets identify discriminative subsequences for classification by searching for patterns that best separate classes in labeled datasets \citep{shapelet}. More recent work leverages scalable motif discovery frameworks, such as STUMPY \citep{law2019stumpy}, which uses efficient matrix \citep{matrix_profile} computations to detect repeated patterns across long sequences, enabling fast retrieval of relevant subsequences in large datasets. \citet{VanWesenbeeck2023LoCoMotif} extend this line of work to local contrastive motif learning, using self-supervised embeddings to identify semantically meaningful motifs across variable-length sequences, even in low-data settings. Despite these advances, existing motif-based methods have not been widely applied to robotics or autonomous driving data, where task-relevant segments are rare, temporally extended, and high-dimensional. By leveraging motif discovery for few-shot retrieval, our approach can efficiently identify meaningful behavioral patterns from minimal demonstrations, providing a scalable and generalizable solution for real-world robotic learning.

\begin{figure}[!h]
    \centering
    \includegraphics[width=\textwidth]{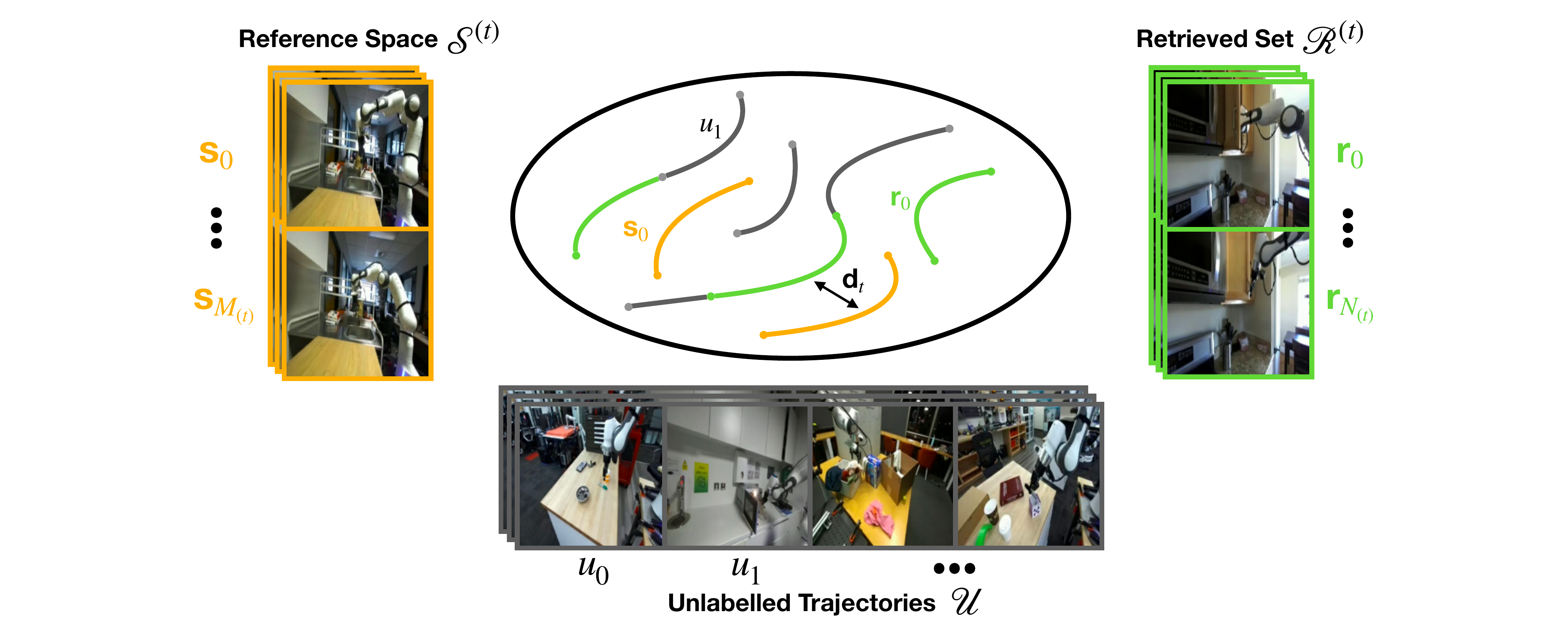}
    \caption{Retrieval Framework. The time-series encoder $f_\theta$ uses labeled dataset $\mathcal{S}$ and create a metric space where similar tasks are grouped together and create prototypes while different tasks are separated from each others. The embeddings of unlabeled trajectories $\mathcal{U}$ are compared with prototypes using learned metric $Dist(.)$ to find the closest match and create retrieved set $\mathcal{R}$.}
    \label{fig:retrieval_framework}
\end{figure}

\section{ROSER}
In this work, we propose a few-shot metric learning framework for retrieving semantically similar robotic and driving maneuvers from long, unsegmented time-series demonstrations. Our approach learns a task-agnostic embedding space in which short temporal sequences corresponding to the same maneuver are clustered together, while dissimilar behaviors are separated. The learned embedding enables scalable retrieval using only a few reference examples per task.

\textbf{Preliminaries} $\mathcal{S}$ denotes the labeled meta-training set used to learn the encoder $f_\theta$. For a target task $t$, $\mathcal{S}^{(t)}$ denotes its support set of size $K_{\text{shot}}$, and $\mathbf{c}^{(t)}$ is the corresponding prototype (Eq.~\ref{eq:prototype}). $\mathcal{U}$ denotes the unlabeled corpus of long unsegmented trajectories from which we retrieve matching subsequences using sliding windows of size $W$ and stride $s$ (see Table~\ref{tab:notation} and Appendix~\ref{app:metrics} for a full notation overview). 

\subsection{Problem Formulation}
Let $\mathcal{S} = \{(\mathbf{X}_i, y_i)\}$ be a dataset of labeled robotic tasks, where $\mathbf{X}_i \in \mathbb{R}^{L_i \times F}$ represents a sequence of robot states (e.g., joint angles, end-effector pose, gripper state, velocity, yaw rate) of length $L_i$, and $y_i \in \{1, \dots, C\}$ is the task label. 

Our goal is to learn an embedding function,
\[
f_\theta : \mathbb{R}^{L \times F} \rightarrow \mathbb{R}^{d_{emb}}
\]
that maps raw robotics time-series segments into a shared metric space. At test time, given a target task $t$ specified by a small support set $\mathcal{S}^{(t)}$ (with $K_{\text{shot}}$ examples), our goal is to retrieve all matching subsequences from a large, unlabeled offline dataset $\mathcal{U} = \{\mathbf{X}_j\}$ where $\mathbf{X}_j \in \mathbb{R}^{L_j \times F}$ and typically $L_j \gg L_i$. Our retrieval framework is illustrated in Figure \ref{fig:retrieval_framework}.

\subsection{Time-Series Encoder}
We adopt a Prototypical Network architecture \citep{snell2017prototypical} to learn a metric space for task retrieval. Unlike standard image-based implementations, our encoder $f_\theta$ must handle variable-length sequences inherent to robotic tasks. 

We use a 1D convolutional network as the sequence encoder to extract time-series data features. Our choice of a streamlined 1D CNN backbone over higher-capacity architectures, such as Transformers \citep{vaswani2023attentionneed} or Large Language Model (LLM) adaptations \citep{Chang2023LLM4TSTF}, is motivated by the extreme data scarcity inherent in few-shot robot imitation. In our setting, $f_\theta$ must learn a discriminative metric space using a minimal support set of reference samples. While Transformers have shown success in large-scale sequence modeling, they possess a "weak" inductive bias, requiring massive datasets to learn temporal dependencies from scratch \citep{dosovitskiy2021an}. In contrast, 1D CNNs provide two critical inductive biases for robotic trajectories, locality and shift-equivariance.

\textbf{Locality:} Robotic control signals are governed by physical laws where the state at time $t$ is most strongly correlated with its immediate neighbors. Convolutional kernels explicitly exploit this local structure.

\textbf{Temporal Shift-Equivariance:} A specific maneuver (e.g., a "grasp") remains semantically identical regardless of when it occurs within the sliding window. CNNs are mathematically equivariant to such shifts, whereas Attention mechanisms must rely on complex positional encodings to learn this invariance \citep{Ismail_Fawaz_2019}.

Furthermore, high-parameter models are prone to manifold collapse and overfitting when the support set is limited to only a few labeled examples \citep{snell2017prototypical}. By utilizing a constrained convolutional architecture, we enforce a smoother latent manifold that generalizes more robustly to the novel task variations encountered in our few-shot experiments. The architecture details of the sequence encoder is reported in Appendix \ref{app:model}.

\subsection{Prototype Construction}
Prototypical  Networks operate by computing a "prototype" $\mathbf{c}^{t}$ for each task $t$, defined as the mean embedding of its support examples \citep{snell2017prototypical}. For a task $t$ with support set $\mathcal{S}^{(t)}$:
\begin{equation}
 \mathbf{c}^{(t)} = \frac{1}{K_{\text{shot}}} \sum_{\mathbf{x} \in \mathcal{S}^{(t)}} f_\theta(\mathbf{x}) 
 \label{eq:prototype}
\end{equation}
where $\mathcal{S}^{(t)}$ is the reference support set sampled within the current training episode. The prototype $\mathbf{c}^{(t)}$ thus represents the task's centroid in the embedding space, acting as the representative template for episodic optimization.

\subsection{Episodic Training Paradigm}
To enable few-shot generalization, we train the network using an episodic paradigm. In each training iteration, we sample an "episode" rather than a standard mini-batch. An episode is constructed by randomly selecting $N_{\text{way}}$ tasks from the task set $\mathcal{T}$. For each selected task $t$, we sample $K_{\text{shot}}$ support examples to form $\mathcal{S}^{(t)}$ and $N_{q}$ query examples to form $\mathcal{Q}^{(t)}$.

The training objective is to minimize the negative log-probability of the true task $t$ for each query example $\mathbf{x} \in \mathcal{Q}^{(t)}$ via a softmax over distances in the latent manifold:
\begin{equation}
 p(y=t|\mathbf{x}) = \frac{\exp\left(-\text{Dist}(f_\theta(\mathbf{x}), \mathbf{c}^{(t)})\right)}{\sum_{t'} \exp\left(-\text{Dist}(f_\theta(\mathbf{x}), \mathbf{c}^{(t')})\right)}   
\end{equation}
where $\text{Dist}(\mathbf{z}, \mathbf{c}^{(t)}) = ||\mathbf{z} - \mathbf{c}^{(t)}||^2$ is the squared Euclidean distance. The loss function $\mathcal{L}$ for an episode is defined as the average negative log-likelihood over all query samples across all sampled tasks:
\begin{equation}
     \mathcal{L} = -\frac{1}{N_{\text{way}} N_{q}} \sum_{t=1}^{N_{\text{way}}} \sum_{\mathbf{x} \in \mathcal{Q}^{(t)}} \log p(y=t|\mathbf{x})
 \end{equation}

This optimization objective forces the encoder $f_\theta$ to cluster embeddings of the same task type tightly around their respective prototypes while maximizing the inter-class distance between different task types.

\begin{algorithm}[!t]
  \caption{Robotic Sequence Retrieval Task (General)}
  \label{alg:task_retrieval}
  \begin{algorithmic}
    \State {\bfseries Input:} Reference support set $\mathcal{S}$, Unlabeled dataset $\mathcal{U}$
    \State {\bfseries Components:} Encoder (or feature map) $f_\theta$, distance $\text{Dist}(\cdot,\cdot)$, template builder $\text{BuildTemplates}(\cdot)$
    \State{\bfseries Hyperparameters:} $W$ (window size), $s$ (stride), $\tau$ (NMS threshold)
    \vskip 0.1in

    \State \textbf{Template Construction:}
    \State Compute templates $\mathcal{C} = \text{BuildTemplates}(\mathcal{S}, f_\theta)$
    \State Initialize retrieval set $\mathcal{R} = \emptyset$

    \For {{\bfseries each} template $\mathbf{c} \in \mathcal{C}$}
        \For{{\bfseries each} trajectory $u \in \mathcal{U}$}
            \For{{\bfseries each} window $\mathbf{w}_t \in u$ with stride $s$}
                \State Compute distance $d_t = \text{Dist}(f_\theta(\mathbf{w}_t), \mathbf{c})$
                \State $\mathcal{R} = \mathcal{R} \cup \{(d_t, \text{start}_t, \text{end}_t)\}$
            \EndFor
        \EndFor
    \EndFor

    \State Sort $\mathcal{R}$ by distance $d_t$ in ascending order
    \State Apply Non-Maximum Suppression (NMS) on $\mathcal{R}$ using threshold $\tau$
    \State {\bfseries Output:} Ranked list of retrieved task candidates
  \end{algorithmic}
\end{algorithm}

\begin{algorithm}[!t]
  \caption{ROSER Episodic Training}
  \label{alg:roser_training}
  \begin{algorithmic}
    \State {\bfseries Input:} Labeled training data $\mathcal{S}$
    \State {\bfseries Hyperparameters:} $N_{\text{way}}, K_{\text{shot}}, N_q$
    \vskip 0.1in

    \Repeat
    \State Sample $N_{\text{way}}$ task classes from $\mathcal{S}$
        \For{{\bfseries each} task $t \in \{1, \dots, N_{\text{way}}\}$}
            \State Sample support set $\mathcal{S}^{(t)}$ ($K_{\text{shot}}$ samples) and query set $\mathcal{Q}^{(t)}$ ($N_q$ samples) from $\mathcal{S}$
            \State Compute prototype $\mathbf{c}^{(t)} = \text{Aggregate}(\{f_\theta(\mathbf{x}) : \mathbf{x} \in \mathcal{S}^{(t)}\})$
        \EndFor
        \State $\mathcal{L} = \sum_{t} \sum_{\mathbf{x} \in \mathcal{Q}^{(t)}} -\log \frac{\exp(-\text{Dist}(f_\theta(\mathbf{x}), \mathbf{c}^{(t)}))}{\sum_{t'} \exp(-\text{Dist}(f_\theta(\mathbf{x}), \mathbf{c}^{(t')}))}$
        \State Update $\theta$ via backpropagation on $\mathcal{L}$
    \Until{convergence}

    \State {\bfseries Output:} Trained encoder $f_\theta$
  \end{algorithmic}
\end{algorithm}

\subsection{Retrieval}
Following training, the encoder $f_\theta$ is deployed for retrieval on unsegmented offline sequences. Task-specific prototypes $\mathbf{c}^{(t)}$ are computed from the $K_{\text{shot}}$-shot support examples for each task following equation \ref{eq:prototype}. Once the prototype is established, a sliding window search is executed across the unlabeled sequences from the dataset $\mathcal{U}$. We extract temporal windows $\mathbf{w}_t$ where the window size $W$ and stride $s$ depends on the sequence length and each window is mapped into the metric space via the trained encoder to compute the squared Euclidean distance $d_t = ||f_\theta(\mathbf{w}_t) - \mathbf{c}^{(t)}||^2$.

To handle the high density of overlapping candidates generated by the sliding window, we incorporate a post-processing stage using Non-Maximum Suppression (NMS) \citep{hosang2017learningnonmaximumsuppression}. By ranking detected windows according to their alignment distance and filtering those that exceed an overlap threshold $\tau$, we ensure that the system retrieves distinct physical maneuvers rather than redundant temporal fragments of the same event. This combined approach allows for precise and scalable identification of target tasks within continuous, high-dimensional robotic data. Finally, the remaining candidates are ranked by their distance $d_t$ in ascending order, and the top-$k$ non-overlapping windows are retained as the definitive retrieved task candidates.

To evaluate the efficacy of our proposed encoder and compare with baselines, we formulate a general task retrieval framework (Algorithm \ref{alg:task_retrieval}) that accommodates both learned representations and classical distance-based measures. We represent all models through a unified retrieval interface defined by an encoder $f_\theta$ and a distance metric $\text{Dist}(\cdot)$. For ROSER, episodic training (Algorithm \ref{alg:roser_training}) optimizes $\theta$ to minimize a discriminative loss $\mathcal{L}_{\text{task}}$. For high-capacity foundation models such as LLMs or time-series foundation models, we bypass training and utilize frozen pre-trained weights to extract features. Finally, for non-parametric baselines like DTW \citep{Meert_DTAIDistance_2020} and STUMPY \citep{law2019stumpy}, $f_\theta$ is an identity mapping, and retrieval is performed via direct sequence alignment.

% \begin{algorithm}[tb]
%   \caption{General Retrieval Framework}
%   \label{alg:task_retrieval}
%   \begin{algorithmic}
%     \STATE {\bfseries Input:} Reference data $\mathcal{S}$, Unlabeled dataset $\mathcal{U}$
%     \STATE {\bfseries Hyperparameters:} $W$ (window size), $s$ (stride), $\tau$ (NMS threshold)
%     \vskip 0.1in
    
%     \STATE \textbf{\textcolor{blue}{/* Phase 2: Task Retrieval (All Baselines) */}}
%     \STATE \textbf{Template:} Compute $\mathcal{C} = f_\theta(x),  x \in \mathcal{S}$ where $f_\theta$ is encoder for LLM or identity mapping for non-parametric model.
%     \STATE Initialize retrieval set $\mathcal{R} = \emptyset$
%     \FOR{{\bfseries each} trajectory $u \in \mathcal{U}$}
%         \FOR {{\bfseries each} $c \in \mathcal{C}$}
%             \FOR{{\bfseries each} window $\mathbf{w}_t \in u$ with stride $s$}
%                 \STATE Compute distance $d_t = \text{Dist}(f_\theta(\mathbf{w}_t), \mathbf{c})$
%                 \STATE $\mathcal{R} = \mathcal{R} \cup \{(d_t, \text{start}_t, \text{end}_t)\}$
%             \ENDFOR
%         \ENDFOR
%     \ENDFOR
%     \STATE Sort $\mathcal{R}$ by distance $d_t$ in ascending order
%     \STATE Apply Non-Maximum Suppression (NMS) on $\mathcal{R}$ using threshold $\tau$
%     \STATE {\bfseries Output:} Ranked list of retrieved task candidates
%   \end{algorithmic}
% \end{algorithm}

\section{Experiments}
%Our experiments address the following questions: (1) Which representation families (LLM embeddings, time-series foundation models, classical matching, or ROSER) best preserve both distributional similarity and temporal dynamics across domains? (2) When do classical matching methods (e.g., DTW, motif matching) succeed or fail compared to learned metric retrieval? (3) How sample-efficient is ROSER with respect to the number of reference demonstrations ($K_{\text{shot}}$)? (4) What is the quality--latency trade-off across methods, and can ROSER achieve sub-millisecond matching without sacrificing retrieval quality? (5) How do retrieval hyperparameters (top-$k$ budget) affect the similarity--diversity trade-off of the retrieved set? (6) Which sensory feature groups are most critical for retrieval in manipulation versus driving, and how robust is ROSER to feature removal?
\subsection{Experimental Setup}
\textbf{Baselines and Ablations.} 
We compare our approach with representative sequence and motif-based retrieval methods such as STUMPY \citep{law2019stumpy, matrix_profile}, Dtaidistance \citep{Meert_DTAIDistance_2020}, and Shapelets \citep{shapelet}. We also evaluate three LLM-based embedding models (Qwen3 \citep{yang2025qwen3technicalreport}, Gemma3 \citep{gemmateam2025gemma3technicalreport}, and LLama-Nemotron \citep{babakhin2025llamaembednemotron8buniversaltextembedding}) as well as one time-series foundation model (MOMENT \citep{goswami2024momentfamilyopentimeseries}).

\textbf{Datasets} We evaluate our approach on three large-scale datasets LIBERO \citep{liu2023libero}, DROID \citep{khazatsky2024droid}, and nuScenes \citep{caesar2020nuscenes}, covering robot manipulation and autonomous driving, focusing on fundamental manipulation tasks and driving maneuvers. Each dataset provides rich multimodal data and supports reproducible evaluation of learning, transfer, and robustness. The details of the datasets are reported in Appendix \ref{app:datasets}.

\textbf{Sliding-window retrieval and post-processing.} Retrieval is performed by extracting fixed-length windows of size $W$ with stride $s$ from each sequence in $\mathcal{U}$ and ranking windows by $\text{Dist}(f_\theta(\mathbf{w}), \mathbf{c}^{(t)})$ (Algorithm~\ref{alg:task_retrieval}). To avoid redundant overlaps, we apply Non-Maximum Suppression (NMS) with threshold $\tau$. The window size $W$ for retrieval is set to the mean length of the support examples, and we use stride $s=W/4$ to ensure dense coverage of potential start times.

\begin{table}[!t]
\caption{Performance of baselines and ROSER on LIBERO dataset. Dark color represent best result and light color represent second best result. All models have same top-k results. Time-series models like Stumpy, Dtaidistance, and ROSER perform better compared to Large Models. ROSER performs best or second best in all of the metrics. }
\label{tab:libero-result}
\begin{center}
\begin{small}
\begin{sc}
\setlength{\tabcolsep}{4pt} 
\resizebox{\textwidth}{!}{%
\begin{tabular}{lcccccc}
\toprule
Model & WD $\left(\downarrow\right)$ & DTW NN $\left(\downarrow\right)$ & Spectral WD $\left(\downarrow\right)$ & Temp Corr. $\left(\uparrow\right)$ & Density $\left(\uparrow\right)$ & Diversity $\left(\downarrow\right)$ \\
\midrule
gemma & 0.17 & 12.98 & 0.0028 & 0.51 & 0.32 & 16.29 \\
llama & 0.16 & 12.08 & 0.0024 & 0.57 & 0.37 & 14.61 \\
momentfm & 0.14 & 11.04 & 0.0023 & 0.62 & 0.41 & 15.10 \\
qwen & 0.17 & 13.04 & 0.0029 & 0.51 & 0.29 & 15.07 \\
shapelet & 0.19 & 11.81 & 0.0034 & 0.56 & 0.40 & 11.74 \\
dtaidistance & 0.092 & \first{6.42} & 0.0022 & 0.70 & \first{0.75} & \first{7.27}\\
stumpy & \second{0.11} & 7.93 & \first{0.0018} & \first{0.72} & 0.64 & 10.92 \\
ROSER & \first{0.086} & \second{6.57} & \second{0.0019} & \first{0.72} & \second{0.69} & \second{8.8} \\
\bottomrule
\end{tabular}%
}
\end{sc}
\end{small}
\end{center}
\end{table}

\begin{table}[!t]
\caption{Performance of baselines and ROSER on nuScenes dataset. Time-series models like Stumpy, Dtaidistance, and ROSER perform better compared to Large Models. ROSER performs best or second best in most of the metrics.}
\label{tab:nuscenes-result}
\begin{center}
\begin{small}
\begin{sc}
\setlength{\tabcolsep}{4pt} 
\resizebox{\textwidth}{!}{%
\begin{tabular}{lcccccc}
\toprule
Model & WD $\left(\downarrow\right)$ & DTW NN $\left(\downarrow\right)$ & Spectral WD $\left(\downarrow\right)$ & Temp Corr. $\left(\uparrow\right)$ & Density $\left(\uparrow\right)$ & Diversity $\left(\downarrow\right)$ \\
\midrule
gemma & 1.02 & 36.56 & 0.00071 & 0.35 & 0.59 & 51.69 \\
llama & 0.73 & 30.74 & 0.00073 & 0.37 & 0.79 & 50.05 \\
momentfm & 0.67 & 24.09 & 0.00086 & 0.42 & 0.79 & 48.86 \\
qwen & 0.96 & 43.95 & 0.00083 & 0.28 & 0.49 & 74.38 \\
shapelet & 0.93 & 32.25 & 0.0015 & 0.14 & 0.46 & 48.82 \\
dtaidistance & 0.54 & \second{13.19} & \second{0.00069} & 0.48 & 1.07 & \first{\textbf{19.98}} \\
stumpy & \second{0.51} & 21.38 & 0.00071 & \second{0.53} & \second{0.85} & 52.70 \\
ROSER & \first{0.27} & \first{12.84} & \first{0.00056} & \first{0.53} & \first{1.30} & \second{26.69} \\
\bottomrule
\end{tabular}%
}
\end{sc}
\end{small}
\end{center}
\end{table}

\begin{table}[!t]
\caption{Performance of baselines and ROSER on DROID dataset. Time-series models like Stumpy, Dtaidistance, and ROSER perform better compared to Large Models. ROSER performs best or second best in most of the metrics.}
\label{tab:droid-result}
\begin{center}
\begin{small}
\begin{sc}
\setlength{\tabcolsep}{4pt} 
\resizebox{\textwidth}{!}{%
\begin{tabular}{lcccccc}
\toprule
Model & WD $\left(\downarrow\right)$ & DTW NN $\left(\downarrow\right)$ & Spectral WD $\left(\downarrow\right)$ & Temp Corr. $\left(\uparrow\right)$ & Density $\left(\uparrow\right)$ & Diversity $\left(\downarrow\right)$ \\
\midrule
Gemma & 0.23 & 30.33 & 0.0021 & 0.35 & 0.69 & 49.54 \\
Llama & 0.23 & 29.56 & 0.0018 & 0.37 & 0.72  & 48.66 \\
Momentfm & 0.23 & 29.88 & 0.0019 & 0.36 & 0.64 & 48.87\\
Qwen & 0.24 & 31.03 & 0.0023 & 0.33 & 0.65 & 50.05 \\
Shapelet & 0.26 & 29.64 & 0.0028 & 0.37 & 0.72 & 46.50 \\
Dtaidistance & 0.21 & \second{25.23} & \first{0.0017} & 0.37 & 0.80 & \first{34.97} \\
Stumpy & \second{0.20} & 28.41 & \second{0.0019} & \first{0.44} & \second{0.82} & 48.17 \\
ROSER & \first{0.16} & \first{22.98} & 0.0021 & \second{0.39} & \first{0.96} & \second{37.00} \\
\bottomrule
\end{tabular}%
}
\end{sc}
\end{small}
\end{center}
\end{table}

\textbf{Implementation Details.} 
We implement a $K$-shot robotics data retrieval model to retrieve data from large unlabeled datasets. Each sample is represented as a multivariate time series. We train for 200 episodes using Adam optimizer \citep{kingma2017adammethodstochasticoptimization} with learning rate $1\mathrm{e}{-3}$ on an NVIDIA A100 GPU.

\textbf{Evaluation Metrics}
\textbf{Primary objective.} Our primary goal is to retrieve task-consistent segments (i.e. semantically correct maneuvers) from long unsegmented logs. Because exhaustive segment-level labels are not always available at scale, we  evaluate retrieval quality using complementary unsupervised proxies: \textit{distributional similarity} measures global statistical alignment between retrieved and reference data; \textit{temporal dynamics} measures preservation of realistic temporal structure and trajectory evolution; and \textit{density and diversity} measure the compactness of retrieved segments and their spread in feature space.
To quantify these properties, we employ a set of low-level alignment metrics such us, Wasserstein Distance (WD), DTW Nearest Neighbor (DTW NN), Spectral Wasserstein Distance (Spectral WD), and high-level behavioral consistency metrics like, Temporal Correlation, Density, and Diversity, capturing both precise trajectory alignment and task-relevant semantic structure. A detailed mathematical formulation for each metric is provided in Appendix \ref{app:metrics}

\subsection{Results}
% \begin{figure*}[t]
%     \centering
%     \includegraphics[width=1.0\textwidth]{figures/retrieval_quality_horizontal.pdf}
%     \caption{nuScenes right turn performance. ROSER retrieves samples close the the target data while Stumpy and Dtaidistance retrieves many left turns with positive yaw rate.}
%     \label{fig:nuscene-right}
% \end{figure*}
\textbf{Which representation families best support robotic sequence retrieval across domains?}  
To understand which representation paradigms are most suitable for robotic sequence retrieval, we compare ROSER against (i) LLM-based embeddings (Llama, Gemma, Qwen), (ii) a time-series foundation model (MomentFM), and (iii) classical time-series matching methods (STUMPY, Dtaidistance, Shapelets) across LIBERO, nuScenes, and DROID. Tables \ref{tab:libero-result}--\ref{tab:droid-result} reveal a consistent pattern: methods that operate directly on time-series structure dominate large language embeddings on both distributional and temporal criteria. In particular, ROSER achieves the most favorable overall profile (best or second-best across nearly all metrics), indicating that a task-adaptive metric space learned from a small support set better captures the task-relevant kinematics than frozen text embeddings. Classical matching baselines remain competitive on specific axes (e.g., strong temporal correlation or spectral alignment), but their performance is less consistent across tasks and datasets, suggesting limited robustness to execution variability and feature relevance. MomentFM generally improves over LLM embeddings, yet still trails the top time-series baselines, supporting the hypothesis that retrieval benefits from representations that are tuned (via episodic training) to the task manifold rather than only pretrained generically.

\textbf{When do classical matching methods fail, and what does ROSER retrieve instead?}  
To probe retrieval failure modes, we compare qualitative and quantitative behavior across tasks with varying execution variability. We find that classical matching approaches (e.g., motif-based STUMPY and DTW-based Dtaidistance) are reliable for highly stereotyped segments, but degrade on tasks where the same semantic maneuver admits multiple valid kinematic realizations (speed profiles, timing, or joint-space detours). In these cases, ROSER is more likely to retrieve behaviorally consistent segments that match the task’s defining dynamics rather than superficial pointwise similarity. For example, LIBERO "bottom drawer open" requires collision-avoiding arm motion; ROSER retrieves trajectories that preserve the longer reach-and-avoid structure, while strong baselines often drift to visually similar but semantically different pick-and-place behaviors (Figure \ref{fig:quality-libero-bottom-drawer-open}). We report qualitative results in detail for all tasks in Appendix \ref{sec:appendix-results}.

\begin{figure}[!t]
    \centering
    \includegraphics[width=\linewidth]{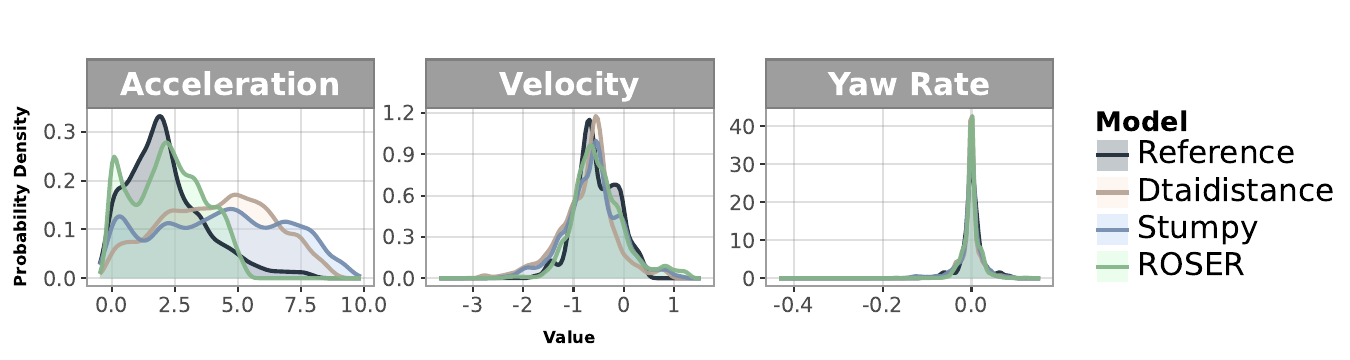}
    \caption{Feature-level distribution visualization for task "regular stop" in the nuScenes benchmark.}
    \label{fig:nuscene-stop}
\end{figure}

This effect is also visible in nuScenes. Table \ref{tab:nuscenes-result} shows that time-series baselines remain competitive on simple maneuvers, but ROSER is substantially more robust for stops and turns where velocity/acceleration/yaw-rate magnitudes and timing vary across scenes. Because ROSER’s metric is shaped by episodic training, it can implicitly reweight features and temporal patterns that are most diagnostic for the queried maneuver. Figure \ref{fig:nuscene-stop} illustrates this: ROSER retrieves "regular stop" segments whose acceleration distribution aligns closely with the reference, while STUMPY and Dtaidistance tend to mix in non-stopping segments that share only partial low-level similarity.

\textbf{How many reference examples are effective for Few-Shot learning?} 
To evaluate the sample efficiency of our few-shot learning approach, we performed an ablation study on the number of reference examples ($K$-shot) used to train the encoder and compute the task prototypes. As detailed in Table \ref{tab:target-size-ablate}, our model demonstrates a high degree of data efficiency, maintaining competitive retrieval performance even when the target set is reduced to 3 or 5 examples. While the best distributional similarity, marked by a Wasserstein Distance of 0.086 on LIBERO and 0.27 on nuScenes, is achieved using 10 reference samples, the performance degradation observed at 7 and 5 samples is minimal. Specifically, the Temporal Correlation and DTW NN metrics remain remarkably stable across all datasets, indicating that the Prototypical Network can reliably define the task manifold even with a sparse support set. However, a slight decline in performance at $K=3$ (e.g., WD increasing to 0.22 in DROID) suggests that a minimum of 5 to 7 samples is the "sweet spot" for balancing minimal human labeling effort with high retrieval accuracy. These findings confirm that our approach effectively bypasses the need for large-scale expert demonstrations, enabling the definition of complex robotic behaviors from a handful of random observations. While more data marginally improves the prototype's stability, the marginal utility of adding samples beyond $K=10$ is expected to diminish, reinforcing the effectiveness of our few-shot formulation for rapid task adaptation.

\textbf{What is the quality-latency trade-off across methods?} Table~\ref{tab:retrieval-speed} reports average per-match retrieval time across datasets. LLM-based embeddings are orders of magnitude slower per comparison, making exhaustive sliding-window retrieval impractical at scale. Classical time-series matching (Dtaidistance, STUMPY) is substantially faster, but still incurs non-trivial per-match cost when the candidate set is large. In contrast, ROSER achieves consistently low latency (sub-millisecond per-match on LIBERO and similarly low overhead on nuScenes and DROID) while also delivering the strongest retrieval quality in Tables \ref{tab:libero-result}--\ref{tab:droid-result}. This combination suggests that a lightweight encoder with a learned metric is a practical design point for large-scale log mining: it enables high-throughput search without relying on heavyweight foundation models.

% ---------------- K-shot ablation ----------------
\begin{table}[!t]
\caption{ROSER performance when number of target samples is varied. We train ROSER in K-3 to K-10 shot manner. Performance degradation is insignificant when we lower K, indicating the robustness of the encoder's ability to learn discriminative features from a few samples.}
\label{tab:target-size-ablate}
\begin{center}
\begin{small}
\begin{sc}
\setlength{\tabcolsep}{4pt} 
\resizebox{\textwidth}{!}{%
\begin{tabular}{lccccccc}
\toprule
Dataset & K-Shot & WD $\left(\downarrow\right)$ & DTW NN $\left(\downarrow\right)$ & Spectral WD $\left(\downarrow\right)$ & Temp Corr. $\left(\uparrow\right)$ & Density $\left(\uparrow\right)$ & Diversity $\left(\downarrow\right)$\\
\midrule
\multirow{4}{*}{Libero}
 & 10 & \first{0.086} & \first{6.57} & \first{0.0019} & \first{0.72} & \first{0.69} & \second{8.8} \\
 & 7  & \second{0.10} & \second{6.84} & \second{0.0022} & \second{0.71} & \second{0.59} & \first{8.38} \\
 & 5  & 0.12 & 8.41 & 0.0026 & 0.68 & 0.49 & 9.61 \\
 & 3  & 0.11 & 7.06 & 0.0025 & 0.69 & 0.55 & 8.72 \\
\midrule
\multirow{4}{*}{nuScenes}
 & 10 & \first{0.27} & \second{12.84} & \second{0.00056} & 0.53 & \second{1.30} & 26.69 \\
 & 7  & 0.35 & \first{12.62} & \first{0.00050} & \first{0.54} & 1.24 & 23.65 \\
 & 5  & \second{0.34} & 12.73 & 0.00070 & \first{0.54} & \second{1.27} & 24.24 \\
 & 3  & 0.36 & 12.92 & 0.00070 & 0.49 & \second{1.27} & \first{22.68} \\
\midrule
\multirow{4}{*}{DROID}
 & 10 & \first{0.16} & \first{22.98} & \second{0.0021} & \second{0.39} & \first{0.96} & \second{37.00} \\
 & 7  & \first{0.16} & 23.96 & \first{0.0019} & 0.37 & \second{0.89} & \first{36.99} \\
 & 5  & \second{0.17} & \second{23.20} & \second{0.0020} & \first{0.40} & 0.88 & 37.50 \\
 & 3  & 0.22 & 25.18 & 0.0022 & 0.38 & 0.72 & 37.75 \\
\bottomrule
\end{tabular}%
}
\end{sc}
\end{small}
\end{center}
\end{table}

\begin{table}[!t]
\vskip 0.1in
\caption{Average retrieval time per match (seconds) across datasets. Each entry reports the mean wall-clock time required to perform a single query–candidate matching operation, measured per offline demonstration and averaged over all retrievals.}
\vskip 0.1in
\label{tab:retrieval-speed}
\begin{center}
\begin{small}
\begin{sc}
\begin{tabular}{lccc}
\toprule
Model & LIBERO $\left(\downarrow\right)$ & nuScenes $\left(\downarrow\right)$ & DROID $\left(\downarrow\right)$ \\
\midrule
Gemma        & 0.1558 & 0.1348 & 0.1347 \\
LLaMA        & 0.8566 & 0.7929 & 0.7919 \\
MomentFM    & 0.1088 & 0.2092 & 0.1040 \\
Qwen         & 0.4449 & 0.4225 & 0.4176 \\
Shapelet     & \second{0.002526} & \first{0.001717} & \first{0.001617} \\
Dtaidistance & 0.09030 & 0.1805 & 0.1770 \\
Stumpy       & 0.009833 & 0.01999 & 0.01818 \\
ROSER        & \first{0.0005163} & \second{0.003873} & \second{0.004549} \\
\bottomrule
\end{tabular}
\end{sc}
\end{small}
\end{center}
\end{table}

\begin{figure}[!t]
    \centering
    \includegraphics[width=0.85\linewidth]{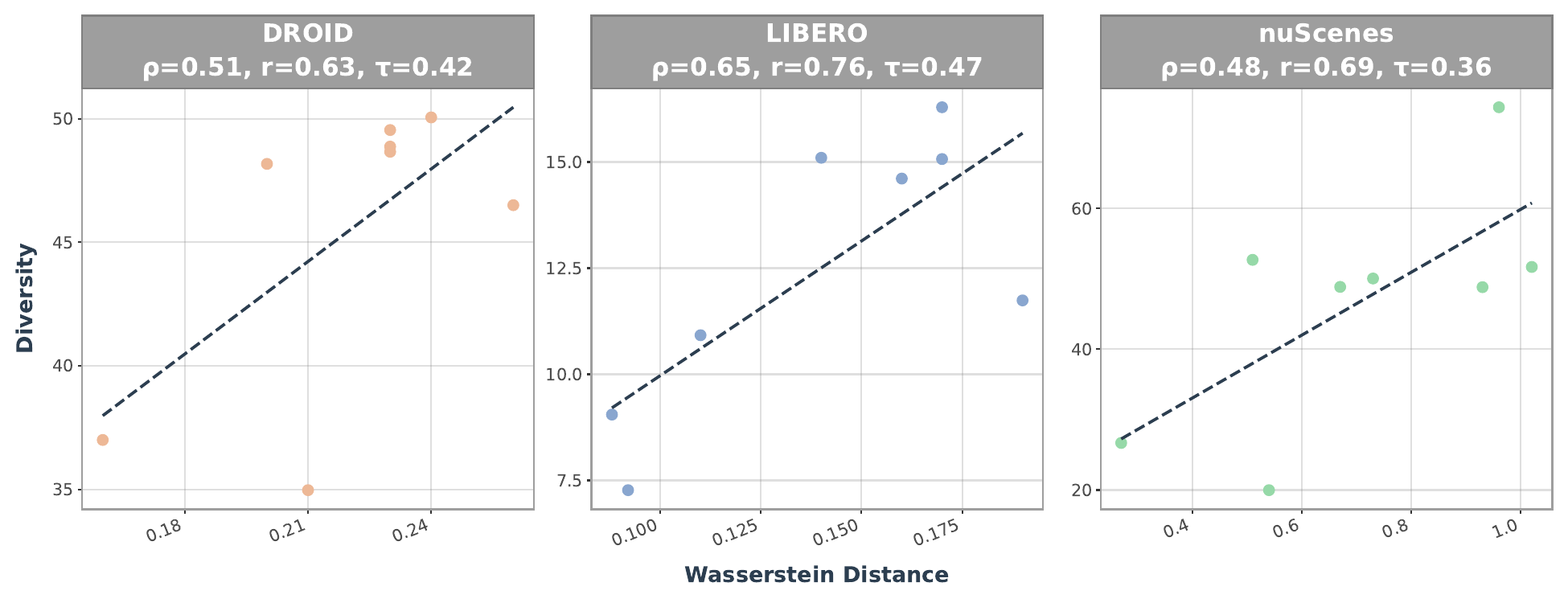}
    \caption{Relationship between Wasserstein Distance and Intra-class distance (diversity). Each point corresponds to a single retrieval model evaluated on the dataset. For each dataset, we report Spearman’s rank correlation ($\rho$), Pearson correlation ($r$), and Kendall Tau ($\tau$).}
    \label{fig:wasserstein-diversity-rel}
\end{figure}

\textbf{What is the trade-off between Diversity and Distributional Similarity?} Figure~\ref{fig:wasserstein-diversity-rel} illustrates a consistent positive correlation between intra-class distance (diversity) and Wasserstein distance across datasets, indicating an inherent trade-off between distributional similarity and behavioral diversity. On DROID, the relationship is moderate (Spearman $\rho=0.51$, Pearson $r=0.63$, Kendall $\tau=0.42$), while LIBERO shows the strongest coupling ($\rho=0.65$, $r=0.76$, $\tau=0.47$), suggesting that in manipulation tasks, closer alignment to the target data distribution is often accompanied by reduced variability among retrieved trajectories. NuScenes exhibits a weaker but still reliable association ($\rho=0.48$, $r=0.69$, $\tau=0.36$), consistent with the higher intrinsic multimodality of navigation behaviors where multiple diverse trajectories may remain distributionally valid. This relationship between diversity and distributional similarity is consistent with previous work in generative model evaluation that sees both as related but separate goals, which often exhibits a trade-off \citep{naeem2020reliablefidelitydiversitymetrics,alaa2022faithfulsyntheticdatasamplelevel,khayatkhoei2023emergentasymmetryprecisionrecall}.
We report the results of our other experiments in Appendix \ref{app:ablation}.

\section{Conclusion}
In this work, we addressed a critical bottleneck in robot learning: the structural incompatibility between how robotic data is collected and how modern learning algorithms consume it. By re-framing data curation as a few-shot retrieval problem, we demonstrated that vast repositories of underutilized robotic logs can be unlocked with minimal supervisio, requiring only a handful of reference examples rather than exhaustive human annotation.
Our contributions are both methodological and empirical. We formalized robotic sequence retrieval as a principled task, introduced ROSER as a lightweight few-shot solution operating in learned metric spaces, and established comprehensive evaluation protocols spanning distributional similarity, temporal dynamics, and diversity-precision trade-offs. Through extensive benchmarking on LIBERO, DROID, and nuScenes, covering manipulation and autonomous driving at scale, we showed that ROSER consistently outperforms classical time-series methods, learned embedding approaches, and large language model baselines in both accuracy and efficiency.
These results have immediate practical implications. ROSER enables researchers to rapidly curate training data for novel tasks by providing few demonstrations, facilitates cross-dataset transfer by identifying analogous behaviors across different embodiments and environments, and supports continual learning by efficiently indexing new data as it becomes available. With sub-millisecond per-match inference and no task-specific training required, ROSER provides a scalable pathway to convert raw robot logs into structured, reusable datasets suitable for imitation learning, policy evaluation, and foundation model pretraining.
Beyond its immediate applications, this work establishes robotic sequence retrieval as a fundamental research problem deserving of sustained attention. As the community continues to scale data collection, through initiatives like the Open X-Embodiment \citep{open_x_embodiment_rt_x_2023}, the ability to efficiently organize, retrieve, and repurpose this data will become increasingly critical. Our benchmark, evaluation protocols, and baseline comparisons provide a foundation for future work in this direction.

\subsection{Limitations and Future Work}

\textbf{Ambiguous transitions and multi-modal behaviors.} Retrieval is most challenging when a target maneuver occurs as a brief transition inside a longer trajectory or when multiple behaviors share similar low-level dynamics (e.g., short decelerations that are not true stops). In these cases, classical matching methods may over-match based on local similarity, while ROSER can still produce false positives when the support set under-specifies the task.

\textbf{Feature dropouts and domain shift.} ROSER is sensitive to missing or corrupted proprioceptive channels and to shifts in sensor scaling across datasets. Our feature ablations (Appendix \ref{app:ablation}) partially quantify this effect; mitigating strategies include per-feature normalization, hard-negative mining, and incorporating additional modalities (e.g., vision) when available.

\textbf{Future Work.} While ROSER demonstrates strong performance on proprioceptive time series, several promising directions remain unexplored. For instance, incorporating visual observations would enable retrieval based on semantic scene content, complementing the kinematic matching our current approach provides. We are actively developing multimodal extensions that fuse trajectory embeddings with vision-language models \citep{radford2021learningtransferablevisualmodels} to support this richer query space.
Finally, by continuously identifying underrepresented behaviors in existing datasets, such systems could prioritize data collection for the most informative regions of the task space, accelerating the path toward generalist robot learning.

\newpage

\bibliography{iclr2026_conference}
\bibliographystyle{iclr2026_conference}

\newpage
\appendix
\section{Appendix}
\label{app:metrics}
\subsection{Evaluation Metrics}
\begin{table}[!ht]
\centering
\small
\renewcommand{\arraystretch}{1.2}

\begin{tabularx}{\columnwidth}{@{} l X @{}} 
\toprule
\textbf{Notation} & \textbf{Description} \\
\midrule
$\mathcal{R}, \mathcal{S}$ & Offline, Retrieved and reference trajectory datasets (collections of tasks) \\
$\mathcal{R}^{(t)}, \mathcal{S}^{(t)}$ & Trajectories for task $t$ \\
$\mathbf{r}_r^{(t)}, \mathbf{s}_i^{(t)}$ & Individual trajectories from task $t$ of $\mathcal{R}^{(t)}$ or $\mathcal{S}^{(t)}$ \\
$L_r^{(t)}, {L_i}^{(t)}$ & Lengths of the r-th retrieved and i-th reference trajectories in task $t$ \\
$F$ & Number of features per time step \\
$r_{r,l,f}^{(t)}, s_{i,l,f}^{(t)}$ & Feature $f$ at time $l$ of trajectory $\mathbf{r}_r^{(t)}$ or $\mathbf{s}_i^{(t)}$ \\
$\mathbf{r}_{r,f}^{(t)}, \mathbf{s}_{i,f}^{(t)}$ & Feature-wise temporal signals (vectors of length $L_r^{(t)}$ or ${L}_i^{(t)}$) \\
$\hat{P}_f^{\mathcal{X}}$ & Mean normalized power spectral density for feature $f$ over set $\mathcal{X}$ \\
$k$ & Number of nearest neighbors for local support radii \\
$\rho_i$ & Distance to the $k$-th nearest neighbor in the reference set\\
$W$ & Window length used to compare time series of unequal length \\
$n$ & Number of sliding windows used in window-based metric computation \\
$\mathcal{W}(\cdot,\cdot)$ & 1-Wasserstein (Earth Mover's) distance \\
$\star$ & Discrete cross-correlation operator \\
$\text{DTW}(\cdot,\cdot)$ & Multivariate Dynamic Time Warping distance \\
$d$ & Arbitrary distance metric between vectors $\in \mathbb{R}^{d}$ \\
\bottomrule
\end{tabularx}
\vskip 0.15in
\caption{Notation used throughout the evaluation metrics}
\label{tab:notation}
\end{table}
\noindent\textbf{Note:} In this appendix, we use $\mathcal{S}$ to denote the \emph{reference} trajectory set for metric computation; this is distinct from the labeled meta-training set $\mathcal{S}$ used in the main method section.

The retrieved and reference datasets may contain trajectories of varying lengths. To enable a consistent and fair evaluation of retrieval quality across these time series, we employ the following window-based metric computation method.

Let the retrieved and reference datasets be collections of tasks:
\[
\mathcal{R} = \{\mathcal{R}^{(t)}\}_{t \in \mathcal{T}}, \quad 
\mathcal{R}^{(t)} = \{\mathbf{r}_r^{(t)} \in \mathbb{R}^{L_r^{(t)} \times F}\}_{r=1}^{N_t},
\]
\[
\mathcal{S} = \{\mathcal{S}^{(t)}\}_{t \in \mathcal{T}}, \quad 
\mathcal{S}^{(t)} = \{\mathbf{s}_i^{(t)} \in \mathbb{R}^{L_i^{(t)} \times F}\}_{i=1}^{M_t}.
\]
where \(\mathcal{T}\) indexes tasks, \(L_r^{(t)}, L_i^{(t)}\) are trajectory lengths, and \(F\) is the feature dimension.  

Define the sliding-window size for each retrieved-reference pair:
\[
W_{ri}^{(t)} = \min(L_r^{(t)}, L_i^{(t)}).
\]
Extract overlapping windows of length \(W_{ri}^{(t)}\) from the longer trajectory. Let \(\mathbf{x}^{(t,k)}\) denote the \(k\)-th window of trajectory \(\mathbf{x}\), and let the number of windows be
\[
n = \max(L_r^{(t)}, L_i^{(t)}) - W_{ri}^{(t)} + 1.
\]
Then, the metric between \(\mathbf{r}_r^{(t)}\) and \(\mathbf{s}_i^{(t)}\) is
\[
\mathcal{M}^{(t)}(\mathbf{r}_r^{(t)}, \mathbf{s}_i^{(t)}) = 
\frac{1}{n} \sum_{k=1}^{n} 
\mathcal{M} \big( \mathbf{r}_r^{(t,k)}, \mathbf{s}_i^{(t,k)} \big),
\]
where if the trajectories are equal length (\(L_r^{(t)} = L_i^{(t)}\)), then \(n=1\) and \(\mathbf{r}_r^{(t,1)} = \mathbf{r}_r^{(t)}\), \(\mathbf{s}_i^{(t,1)} = \mathbf{s}_i^{(t)}\).  

\subsubsection{Distributional Similarity}

Distributional metrics assess whether the retrieved trajectories match the marginal feature distributions of the reference set, regardless of temporal ordering. 

\textbf{Wasserstein Distance.}  
For task $t$, collect feature values across all trajectories and timesteps:
\[
R_f^{(t)} = \{ r_{r,l,f}^{(t)} \mid \mathbf{r}_r^{(t)} \in \mathcal{R}^{(t)},\, l = 1 \ldots L_r^{(t)} \}
\]
\[
S_f^{(t)} = \{ s_{i,l,f}^{(t)} \mid \mathbf{s}_i^{(t)} \in \mathcal{S}^{(t)},\, l = 1 \ldots L_i^{(t)} \}
\]
The 1D Wasserstein distance for feature $f$ is computed as
\[
\mathcal{W}(R_f^{(t)}, S_f^{(t)}) 
= \int_{-\infty}^{\infty} \big| F_{R_f^{(t)}}(x) - F_{S_f^{(t)}}(x) \big| \, dx
\approx \sum_{i=1}^{n} \big| \text{CDF}_{R_f^{(t)}}(x_i) - \text{CDF}_{S_f^{(t)}}(x_i) \big| \, \Delta x_i
\]
and the per-task feature-wise distance is
\[
d(\mathcal{R}^{(t)}, \mathcal{S}^{(t)}) = \frac{1}{F} \sum_{f=1}^{F} \mathcal{W}(R_f^{(t)}, S_f^{(t)}).
\]
\subsubsection{Temporal Dynamics}
Temporal metrics measure whether retrieved trajectories exhibit realistic evolution over time, capturing alignment, phase shifts, and sequential correlations.

\textbf{Nearest-Neighbor Dynamic Time Warping (DTW).}  
For task $t$, each retrieved trajectory is aligned to its most similar reference trajectory:
\[
\text{DTW}_{\text{NN}}(\mathcal{R}^{(t)}, \mathcal{S}^{(t)}) =
\frac{1}{|\mathcal{R}^{(t)}|} \sum_{\mathbf{r}_r^{(t)} \in \mathcal{R}^{(t)}} 
\min_{\mathbf{s}_i^{(t)} \in \mathcal{S}^{(t)}} \text{DTW}(\mathbf{r}_r^{(t)}, \mathbf{s}_i^{(t)}).
\]
Sliding windows are applied when $L_r^{(t)} \neq L_i^{(t)}$, averaging DTW over all windows.

\textbf{Temporal Cross-Correlation (TCC).}  
For task $t$, feature-wise signals are cross-correlated:
\[
\text{TCC}^{(t)} =
\frac{1}{F\,|\mathcal{R}^{(t)}|\,|\mathcal{S}^{(t)}|}
\sum_{f=1}^{F} \sum_{\mathbf{r}_r^{(t)} \in \mathcal{R}^{(t)}} \sum_{\mathbf{s}_i^{(t)} \in \mathcal{S}^{(t)}} 
\frac{\max_\tau (\mathbf{r}_{r,f}^{(t)} \star \mathbf{s}_{i,f}^{(t)})(\tau)}
{\| \mathbf{r}_{r,f}^{(t)} \|_2 \, \| \mathbf{s}_{i,f}^{(t)} \|_2}.
\]
Cross-correlation is computed over windows if lengths differ.  

\textbf{Power Spectral Density (Wasserstein-based) Distance.}  
To compare the frequency content of two datasets for task $t$, we define:
\[
\text{PSD}(\mathcal{R}^{(t)}, \mathcal{S}^{(t)}) =
\frac{1}{F} \sum_{f=1}^{F} \mathcal{W}\big( \hat{P}_f^{\mathcal{R}^{(t)}}, \hat{P}_f^{\mathcal{S}^{(t)}} \big),
\]
where the per-feature PSDs are:
\[
\hat{P}_f^{\mathcal{R}^{(t)}} =
\frac{1}{|\mathcal{R}^{(t)}|} \sum_{\mathbf{r} \in \mathcal{R}^{(t)}} 
\frac{ \big| \mathrm{FFT}_W(\mathbf{r}_f) \big|^2 }
     { \sum_{k=1}^{W} \big| \mathrm{FFT}_W(\mathbf{r}_f)[k] \big|^2 }
\]
\[
\hat{P}_f^{\mathcal{S}^{(t)}} =
\frac{1}{|\mathcal{S}^{(t)}|} \sum_{\mathbf{s} \in \mathcal{S}^{(t)}} 
\frac{ \big| \mathrm{FFT}_W(\mathbf{s}_f) \big|^2 }
     { \sum_{k=1}^{W} \big| \mathrm{FFT}_W(\mathbf{s}_f)[k] \big|^2 }
\]
FFT resolution determined by $W$.

\subsubsection{Density and Diversity}

Density measures how many retrieved trajectories lie near the reference set, while diversity captures how different retrieved trajectories are from one another.

\textbf{k-NN-Based Density \citep{naeem2020reliablefidelitydiversitymetrics}.}  
For task $t$, flatten trajectories or windowed subsequences:
\[
\tilde{\mathbf{r}}_r^{(t)}, \tilde{\mathbf{s}}_i^{(t)} \in \mathbb{R}^{W \cdot F}, \quad 
W = \min(L_r^{(t)},L_i^{(t)}).
\]
Let $d(\cdot,\cdot)$ be a distance metric. Define the k-NN radius for each reference trajectory:
\[
\rho_i^{(t)} := \text{distance to its $k$-th nearest neighbor in } 
\mathcal{S}^{(t)} \setminus \{\mathbf{s}_i^{(t)}\}.
\]
Density is then defined as
\[
\text{Density}^{(t)} =
\frac{1}{|\mathcal{R}^{(t)}|} \sum_{i=1}^{|\mathcal{R}^{(t)}|}
\frac{1}{k} \sum_{j=1}^{|\mathcal{S}^{(t)}|}
\mathbf{1}\{ d(\tilde{\mathbf{r}}_r^{(t)}, \tilde{\mathbf{s}}_i^{(t)}) \le \rho_i^{(t)} \}.
\]
Neighborhood size $k$ can be used as a hyperparameter.

\textbf{Intra-class Distance (ICD).}  
Internal diversity for task $t$ is measured via
\[
\text{ICD}^{(t)} =
\frac{2}{|\mathcal{R}^{(t)}| (|\mathcal{R}^{(t)}| - 1)}
\sum_{i<j} \text{DTW}(\mathbf{r}_r^{(t)}, \mathbf{r}_i^{(t)}).
\]
Sliding windows are applied if sequences differ in length.  

\subsection{Baseline Implementations}
\label{app:baselines}

This appendix rigorously details the baseline retrieval methods evaluated in our experiments. For each benchmark, we fix a retrieval budget $k$, corresponding to the number of subsequences retrieved per task: $k{=}150$ for LIBERO, $k{=}50$ for nuScenes, and $k{=}80$ for DROID. Given a reference trajectory segment, each baseline retrieves the top-$k$ most similar subsequences from an offline dataset. We group baselines into (i) LLM-based embedding methods, (ii) foundation time-series models, and (ii) classical matching-based time-series methods.

For a given task $t \in \mathcal{T}$, let's assume one reference sample from $\mathcal{S}$ is:
\[
\mathbf{s}_i^{(t)} \in \mathbb{R}^{L_i^{(t)} \times F}
\]
where \(L_i\) denotes the sequence length and \(F\) the number of features. Now, the offline dataset is
\[
\mathcal{U} = \{\mathbf{u}_j \in \mathbb{R}^{L_j\times F}\}_{j=1}^{N},
\qquad L_j >> L_i^{(t)} \ \forall i,
\]
from which fixed-length candidate subsequences are created using sliding window technique.

\subsubsection{LLM-based Embedding Baselines}
These baselines convert multivariate time-series subsequences into textual representations and apply pretrained language embedding models for similarity-based retrieval. The following embedding models were used as baselines for this paper:
\begin{itemize}
    \item \textbf{Qwen3 Embedding (0.6B)} \citep{yang2025qwen3technicalreport}
    \item \textbf{LLaMA Nemotron Embed (1B)} \citep{babakhin2025llamaembednemotron8buniversaltextembedding}
    \item \textbf{Gemma3 Embedding (0.3B)} \citep{gemmateam2025gemma3technicalreport}
\end{itemize}

\paragraph{Symbolic Encoding via SAX.}
Consider a candidate subsequence of length $L$, $u \in \mathcal{U}$.

Each feature dimension \(f \in \{1,\dots,F\}\) is independently normalized over time:
\[
\tilde{u}_{l,f} = \frac{u_{l,f} - \mu_f}{\sigma_f},
\qquad l = 1,\dots,L,
\]
where
\[
\mu_f = \frac{1}{L} \sum_{l=1}^{L} u_{l,f},
\quad
\sigma_f^2 = \frac{1}{L} \sum_{l=1}^{L} (u_{l,f} - \mu_f)^2.
\]
If \(\sigma_f = 0\), we set \(\tilde{x}_{l,f} = 0\) for all \(l\).

The normalized sequence is discretized using a SAX-inspired symbolic encoding \citep{lin2003symbolic}.
Let \(B\) denote the number of bins (we use \(B=26\)).
Unlike classical SAX, which applies Piecewise Aggregate Approximation and fixed Gaussian breakpoints, we discretize each subsequence using per-subsequence empirical quantiles:
\[
\mathcal{E}_f = \left\{ q_f\!\left(\frac{k}{B}\right) \right\}_{k=0}^{B},
\]
where \(q_f(\cdot)\) is the quantile function of \(\{\tilde{u}_{l,f}\}_{l=1}^{L}\).

Each value \(\tilde{u}_{l,f}\) is mapped to a symbol
\[
s_{l,f} \in \mathcal{A} = \{\texttt{A},\texttt{B},\dots,\texttt{Z}\}, \quad
s_{l,f} = \mathcal{A}_k \;\; \text{if} \;\; \tilde{u}_{l,f} \in [\mathcal{E}_{f,k}, \mathcal{E}_{f,k+1}), 
\quad k \in \{0,\dots,25\}.
\]

This yields, for each feature \(f\), an ordered symbolic string \(\{s_{l,f}\}_{l=1}^{L}\).
The final textual representation concatenates the \(F\) feature-wise strings while preserving temporal order.

\paragraph{Prompt Template for LLM-based Embedding.}
Each multivariate time series is converted into a structured textual prompt after SAX discretization.
For each feature dimension \(f \in \{1,\dots,F\}\), an ordered symbolic string
\[
\mathbf{s}_f = \{ s_{l,f} \}_{l=1}^{L}
\]
is produced and formatted as a single line in the prompt. 

The resulting prompt template was used:
\[
\begin{aligned}
\texttt{<feature 1>} &\;[\text{SAX Representation}]:\; \mathbf{s}_{1} \\
\texttt{<feature 2>} &\;[\text{SAX Representation}]:\; \mathbf{s}_{2} \\
&\;\vdots \\
\texttt{<feature F>} &\;[\text{SAX Representation}]:\; \mathbf{s}_{F}
\end{aligned}
\]

where "\texttt{<feature f>}" is the f-th feature of the multivariate time series.

% \paragraph{Sliding-window Retrieval with LLM Embeddings.}
% Candidate subsequences are extracted using a sliding-window strategy.
% For a sequence of length \(L\) and window size \(W\), the number of possible windows is
% \[
% N_w = L - W + 1.
% \]
% Given a maximum window cap \(N_{\max}\), the stride is chosen as
% \[
% \Delta = \left\lceil \max\!\left(1, \frac{N_w}{N_{\max}} \right) \right\rceil.
% \]
% This induces a set of window start indices
% \[
% \mathcal{I} = \{0, \Delta, 2\Delta, \dots\} \cap [0, N_w - 1],
% \qquad |\mathcal{I}| \le N_{\max}.
% \]
\paragraph{LLM Inference.}
If we use a sliding window of size $W$, one candidate window from an unlabeled sample $u$ is $u_{l:l+W}$ where $l$ is the initial index and $l<L_j$.

For simplicity, from now on, let's assume, each reference trajectory $x = \textbf{s}_i^{(t)}$. We define the reference embedding representation as
\[
\mathbf{z_x} = f_{\mathrm{LLM}}\!\left(\text{Prompt}(x\right))
\in \mathbb{R}^E,
\]
which encodes the aggregated textual representation of the reference.

Similarly, for simplicity, let's assume each candidate window is $y = u_{l:l+W}$. It's embedding is:
\[
\mathbf{z}_y = f_{\mathrm{LLM}}\!\left(\text{Prompt}(y)\right)
\in \mathbb{R}^E.
\]

Similarity is measured using cosine distance:
\[
d_{l} = 1 -
\frac{\mathbf{z_x}^\top \mathbf{z}_y}{\|\mathbf{z_x}\|_2 \, \|\mathbf{z}_y\|_2}.
\]
We select the top-$k$ candidates with lowest $d_l$.

\subsubsection{MOMENT-FM.}
MOMENT \citep{goswami2024momentfamilyopentimeseries} is a pretrained foundation model for multivariate time-series representation learning. The reference $x$ is embedded once:
\[
\mathbf{z}_x = f_{\text{MOMENT}}(x),
\]
and each candidate window $y$ is embedded independently. 
\[
\mathbf{z}_y = f_{\text{MOMENT}}(y).
\]
Retrieval is performed using cosine similarity,
\[
d_l = 1 - \frac{\mathbf{z_x}^\top \mathbf{z}_y}{\|\mathbf{z_x}\|_2 \, \|\mathbf{z}_y\|_2},
\]
We select the top-$k$ candidates with lowest $d_l$.

\subsubsection{Matching-based Time-Series Baselines.}
\paragraph{STUMPY (Matrix Profile Matching).}
STUMPY computes the matrix profile, which stores the minimum z-normalized Euclidean distance between a query and all subsequences of equal length \citep{matrix_profile,law2019stumpy}. For multivariate inputs, we compute the z-normalized Euclidean distance jointly across all feature dimensions. We use the implementation \footnote{\url{https://stumpy.readthedocs.io/en/latest/Tutorial_Pattern_Matching.html}}
 from \citep{law2019stumpy}.

For a reference trajectory $x$ and a candidate full trajectory $y$, the distance is
\[
d_i = \left\| \frac{x - \boldsymbol{\mu}_x}{\boldsymbol{\sigma}_x}
       - \frac{y - \boldsymbol{\mu}_y}{\boldsymbol{\sigma}_y} \right\|_2,
\]
where $\boldsymbol{\mu}_x$ and $\boldsymbol{\sigma}_x$ are the mean and standard deviation of $x$ along the feature dimension, and $\boldsymbol{\mu}_y$ and $\boldsymbol{\sigma}_y$ are the mean and standard deviation of $y$ along the feature dimension.

We select the top-$k$ candidates with lowest $d_l$.

\paragraph{Dynamic Time Warping (DTW).}
DTW aligns sequences under non-linear temporal distortions \citep{sako78dynamic}. We use the implementation \footnote{\url{https://dtaidistance.readthedocs.io/en/latest/usage/subsequence.html}} from \citep{Meert_DTAIDistance_2020}.
For a reference trajectory segment $x$ and a candidate window $y$, DTW computes
\[
d_l = \mathrm{DTW}(x,y) 
     = \min_{\pi} \sum_{(n,m)\in \pi} \|x_n - y_m\|_2,
\]
where $n = 1 \ldots L_i^{(t)}$, $m = 1 \ldots L_j$, and $\pi$ is a valid warping path.

We select the top-$k$ candidates with lowest $d_l$.

\paragraph{Shapelet-based Matching.}
Shapelets are discriminative subsequences originally optimized for time-series classification \citep{ye2009shapelets}. We use a pretrained shapelet-based classifier to score fixed-length candidate windows. Each window $y$ is evaluated using the classifier’s predicted class probabilities
\[
p_l = \max_c P(c \mid y),
\]
which serves as a confidence-based retrieval score. We define the retrieval cost as
\[
d_l = -p_l,
\]
We select the top-$k$ candidates with lowest $d_l$.

\subsection{Sequence Encoder}
\label{app:model}
The embedding function $f_\theta$ is implemented as a 1D Convolutional Neural Network (CNN) with $L$ sequential convolutional blocks. Each block consists of a 1D convolution, optional batch normalization, and a non-linear activation (ReLU). For an input sequence $\mathbf{X} \in \mathbb{R}^{L \times F}$, the network computes a sequence of hidden representations through $N_l$ blocks:
\begin{align}
    \mathbf{h}_1 &= \text{ReLU}(\text{BN}(\text{Conv}_{\text{1D}}(\mathbf{X}))) \\
    \mathbf{h}_l &= \text{ReLU}(\text{BN}(\text{Conv}_{\text{1D}}(\mathbf{h}_{l-1}))) \quad \text{for } l=2, \dots, N_l
\end{align}
To aggregate these local temporal features into a fixed-size vector representation, we apply Global Average Pooling (GAP) over the temporal dimension. To handle sequences of variable length, a binary mask is applied so that padded timesteps do not affect the global-pooled embedding.
\begin{equation}
    \mathbf{z} = f_\theta(\mathbf{X}) = \frac{1}{L'} \sum_{t=1}^{L'} \mathbf{h}_{N_l} \in \mathbb{R}^{d_{emb}}
\end{equation}
where $L'$ is the temporal resolution of the final feature map and average is computed only over valid (non-padded) timesteps. This ensures invariance to sequence length while preserving salient temporal patterns.  

\subsection{Datasets}
We explain all 3 datasets we used in this study in details here.
\label{app:datasets}

\textbf{LIBERO \citep{liu2023libero}:} LIBERO (LIfelong learning BEchmark on RObot manipulation) is a simulation-based benchmark for studying lifelong learning, knowledge transfer, and multi-task policy generalization in robot manipulation. We focus on a subset of fundamental tasks, grouped as: \textit{Bottom Drawer Open/Close, Top Drawer Open/Close, Microwave Open/Close, Stove On/Off, and Pick-and-Place}. The benchmark contains 130 language-conditioned tasks organized into suites, including LIBERO-90 for pretraining and LIBERO-10 for downstream evaluation. High-quality human teleoperation demonstrations are provided for all tasks, enabling sample-efficient policy training and rigorous comparison of continual learning approaches. Dataset contains end effector (EE) pose, gripper states, robot joint states and environment videos. We use EE pose, gripper states, and joint states in this study.
    %LIBERO’s procedural generation pipeline allows scalable creation of new tasks, supporting research in generalization, robustness, and task transfer.

\textbf{DROID \citep{khazatsky2024droid}:} DROID (Distributed Robot Interaction Dataset) is a large-scale real-world robot manipulation dataset collected across hundreds of diverse indoor environments. We evaluate fundamental manipulation tasks, including: \textit{Close/Open Cabinet, Close Drawer, Pick-and-Place, and Turn Faucet/Knobs}. The dataset contains tens of thousands of human teleoperation demonstration trajectories with synchronized multi-view RGB video streams, depth, robot proprioceptive measurements (joint positions and velocities), end-effector poses, and optional natural language annotations. DROID provides extensive real-world variation and metadata, supporting training and evaluation of policies that generalize across environments and tasks, and includes released hardware details and code for reproducible data collection and policy development. In this study, we use EE pose, gripper states and joint states similar as LIBERO. We randomly sample around 500 relevant demonstrations using the provided language instructions for this study.

\textbf{nuScenes \citep{caesar2020nuscenes}:} nuScenes is a large-scale multimodal dataset for autonomous driving research, collected using a full sensor suite on autonomous vehicles. We focus on fundamental driving maneuvers, including: \textit{Left Turn, Right Turn, Straight Driving, and Regular Stop}. The dataset provides synchronized data from six cameras, five radars, and one LiDAR sensor, offering 360° coverage of each scene. Each scene spans ~20 seconds and is densely annotated with 3D bounding boxes for 23 object classes and attributes, along with object trajectories, motion information, and contextual metadata. In this study, we only use dynamics data like forward velocity, forward acceleration, and yaw rate.
%nuScenes supports a wide range of perception and prediction tasks, including 3D object detection, tracking, semantic segmentation, and motion forecasting, and provides standardized evaluation metrics and benchmarks for reproducible comparison of autonomous driving models.

\section{Ablation Studies}
In this section, we explain in details the rest of our experiments to validate ROSER. We ablated the number of retrieved top-$k$ and the proprioceptive features.
\label{app:ablation}
%\textbf{Feature ablation.} Table~\ref{tab:ablate-dimension} shows that ROSER’s retrieval quality is driven by task-critical state variables, with performance degrading in a manner consistent with domain semantics. In manipulation datasets (LIBERO and DROID), removing end-effector pose or joint states leads to clear increases in Wasserstein distance, DTW nearest-neighbor distance, and diversity, together with reduced temporal correlation and density. This indicates that accurate geometric and kinematic information is essential for preserving both distributional alignment and temporal coherence. In contrast, removing gripper states has a smaller impact, suggesting that global pose and joint configuration dominate retrieval quality. For navigation in nuScenes, motion-related features are most influential: removing velocity or acceleration produces the largest degradation, while yaw rate has a comparatively limited effect. These results confirm that ROSER adapts to the dominant dynamics of each domain and relies primarily on features that encode task-defining structure.

\textbf{What is the impact of the retrieval threshold top-$(k)$ on the quality of the retrieved data distribution?} 
We study the effect of the retrieval threshold $k$ by sweeping the number of retrieved trajectories on LIBERO, nuScenes, and DROID datasets. Trajectories are ranked by Euclidean distance in the learned embedding space to task prototypes, and the top-$k$ samples are selected. Table~\ref{tab:roser_k_sweep} reveals a consistent trade-off between distributional similarity, temporal alignment, and set-level properties. Across all datasets, smaller $k$ yields stronger \emph{distributional similarity}, as indicated by lower Wasserstein Distance (WD). At the same time, \emph{temporal alignment} is highest at low $k$, reflected by lower DTW Nearest Neighbor distance, lower Spectral WD, and higher Temporal Correlation. In contrast, increasing $k$ improves the \emph{set-level properties} of the retrieved collection: Density increases and Diversity improves (lower ICD), indicating broader coverage of task variations. Concretely, the best distributional similarity and temporal alignment are achieved at $k{=}50$ for Libero, $k{=}20$ for nuScenes, and $k{=}40$ for DROID. Beyond these points, WD, DTW NN, Spectral WD, and Temporal Correlation degrade monotonically, while Density and Diversity improve. These results show that Euclidean distance to the task prototype provides a reliable ranking of trajectory relevance, and that low-$k$ retrieval is preferable when preserving distributional similarity and temporal structure is critical, whereas larger $k$ favors coverage and diversity.

\textbf{Which sensory features are critical for cross-dataset task retrieval, and how do feature dependencies vary between manipulation and navigation domains?}  
To identify the most informative features for task-relevant retrieval, we conducted a feature ablation study across all datasets, removing one category of sensory input at a time. As shown in Table \ref{tab:ablate-dimension}, the manipulation-heavy datasets, LIBERO and DROID, exhibit remarkably similar behavior, where joint states and end-effector (EE) position serve as the primary pillars of the task manifold. For both datasets, the exclusion of Joint States led to the most substantial performance drops in DTW NN and distributional Density, confirming that the high-dimensional articulation of the arm is essential for disambiguating complex manipulation tasks. Besides, the exclusion of soint states and end-effector position led to the substantial drop in Wasserstein Distance. Interestingly, the removal of Joint States also led to an artificial inflation of the Diversity metric (e.g., from 8.8 to 12.13 in LIBERO), suggesting that without joint information, the retrieval results become more chaotic and less consistent with the prototype. In contrast, for the nuScenes navigation dataset, the model is most sensitive to Velocity, reflecting the fundamental importance of speed profiles in characterizing driving behaviors like turns or braking. Across all domains, we find that while the model is robust to the removal of secondary features like gripper states or yaw rate, the "kinematic backbone", joint states for robots and velocity for vehicles—is indispensable. This cross-dataset consistency between LIBERO and DROID validates that ROSER successfully identifies a universal manipulation embedding that prioritizes arm configuration over binary gripper states.

% ---------------- Top-K sweep ----------------

\begin{table}[!t]
\caption{ROSER Metrics for Varying Top-$k$. We retrieve a range of samples for each task. Performance is better when the retrieval number is small. Our assumption is tasks are imbalanced and retrieving more samples may include false positives.}
\label{tab:roser_k_sweep}
\begin{center}
\begin{small}
\begin{sc}
\setlength{\tabcolsep}{4pt} 
\resizebox{\textwidth}{!}{%
\begin{tabular}{lccccccc}
\toprule
Dataset & Top-$k$ & WD $\left(\downarrow\right)$ & DTW NN $\left(\downarrow\right)$ & Spectral WD $\left(\downarrow\right)$ & Temp Corr. $\left(\uparrow\right)$ & Density $\left(\uparrow\right)$ & Diversity $\left(\downarrow\right)$ \\
\midrule
\multirow{6}{*}{Libero}
 & 50  & \first{0.080} & \first{4.96} & \first{0.0017} & \first{0.78} & \first{0.82} & \first{6.15} \\
 & 70  & \second{0.081} & \second{5.27} & \second{0.0018} & \second{0.77} & \second{0.77} & \second{6.75} \\
 & 90  & 0.083 & 5.70 & 0.0019 & 0.75 & 0.71 & 7.46 \\
 & 110 & 0.085 & 6.09 & 0.0019 & 0.74 & 0.67 & 8.11 \\
 & 130 & 0.086 & 6.36 & 0.0020 & 0.73 & 0.64 & 8.62 \\
 & 150 & 0.086 & 6.57 & 0.0019 & 0.72 & 0.69 & 8.8 \\
\midrule
\multirow{4}{*}{nuScenes}
 & 20 & \first{0.25} & \first{11.29} & \second{0.00052} & \first{0.57} & \first{1.44} & \first{22.22} \\
 & 30 & \first{0.25} & \second{11.54} & \first{0.00050} & \second{0.56} & \second{1.40} & \second{23.68} \\
 & 40 & \second{0.27} & 12.40 & 0.00054 & 0.54 & 1.32 & 25.60 \\
 & 50 & \second{0.27} & 12.84 & 0.00056 & 0.53 & 1.30 & 26.69 \\
\midrule
\multirow{5}{*}{DROID}
 & 40 & \first{0.16} & \second{22.81} & \first{0.0020} & \second{0.39} & \first{0.98} & \second{36.85} \\
 & 50 & \first{0.16} & \first{22.74} & \first{0.0020} & \first{0.40} & \second{0.97} & \first{36.64} \\
 & 60 & \first{0.16} & 22.87 & \second{0.0021} & \first{0.40} & \second{0.97} & \second{36.74} \\
 & 70 & \first{0.16} & 23.11 & \second{0.0021} & \second{0.39} & 0.95 & 37.03 \\
 & 80 & \first{0.16} & \second{22.98} & \second{0.0021} & \second{0.39} & \second{0.96} & 37.00 \\
\bottomrule
\end{tabular}%
}
\end{sc}
\end{small}
\end{center}
\end{table}

\begin{table}[!t]
\caption{ROSER performance when one of the features is removed. For manipulation tasks, end-effector pose and joint states are the most important. For navigation, velocity and acceleration play major roles.}
\vskip 0.1in
\label{tab:ablate-dimension}
\begin{center}
\begin{small}
\begin{sc}
\setlength{\tabcolsep}{4pt} 
\resizebox{\textwidth}{!}{%
\begin{tabular}{lccccccc}
\toprule
Dataset & Removed Feature & WD $\left(\downarrow\right)$ & DTW NN $\left(\downarrow\right)$ & Spectral WD $\left(\downarrow\right)$ & Temp Corr. $\left(\uparrow\right)$ & Density $\left(\uparrow\right)$ & Diversity $\left(\downarrow\right)$ \\
\midrule
\multirow{4}{*}{Libero}
 & N/A & \first{0.086} & \first{6.57} & \first{0.0019} & \first{0.72} & \first{0.69} & \first{8.8}\\
 & EE Pos & 0.12 & 7.9 & 0.0023 & 0.67 & 0.51 & 9.2 \\
 & Gripper States & \second{0.10} & \second{7.52} & 0.0023 & 0.67 & \second{0.56} & \second{9.12}\\
 & Joint States & 0.12 & 9.28 & \second{0.0022} & \second{0.69} & 0.53 & 12.13 \\
\midrule
\multirow{4}{*}{nuScenes}
 & N/A & \first{0.27} & \first{12.84} & \second{0.00056} & \second{0.53} & \second{1.30} & \second{26.69}\\
 & Velocity & 0.41 & 17.22 & \first{0.0005} & \first{0.54} & 0.95 & 40.7 \\
 & Acceleration & 0.35 & 17.43 & 0.0006 & 0.48 & 1.14 & 32.72 \\
 & Yaw Rate & \second{0.28} & \second{13.76} & 0.0007 & 0.43 & \first{1.34} & \first{26.5} \\
\midrule
\multirow{4}{*}{DROID}
 & N/A & \first{0.16} & \second{22.98} & \second{0.0021} & \second{0.39} & \first{0.96} & \first{37.00}\\
 & EE Pos & 0.21 & 24.49 & \second{0.0021} & \second{0.39} & 0.74 & 42.15\\
 & Gripper States & \first{0.16} & \first{22.93} & \first{0.0019} & \first{0.40} & \second{0.86} & \second{38.28}\\
 & Joint States & 0.22 & 24.89 & 0.0025 & \second{0.39} & 0.76 & 38.42\\
\bottomrule
\end{tabular}%
}
\end{sc}
\end{small}
\end{center}
\end{table}
\clearpage

\newpage
\section{Distribution and Qualitative Results}
\label{sec:appendix-results}

This appendix reports qualitative results for all tasks evaluated across the benchmarks. For each task or maneuver, we present representative demonstrations together with feature-level distribution plots comparing the top-performing methods: Dtaidistance, STUMP, and ROSER (ours).

% ================= LIBERO BENCHMARK =================
\subsection{LIBERO Benchmark}
\label{sec:appendix-libero}

% --- Microwave Open ---
\subsubsection{Microwave Open}
\label{sec:libero-microwave-open}
\begin{figure}[!ht]
    \centering
    \includegraphics[width=\linewidth]{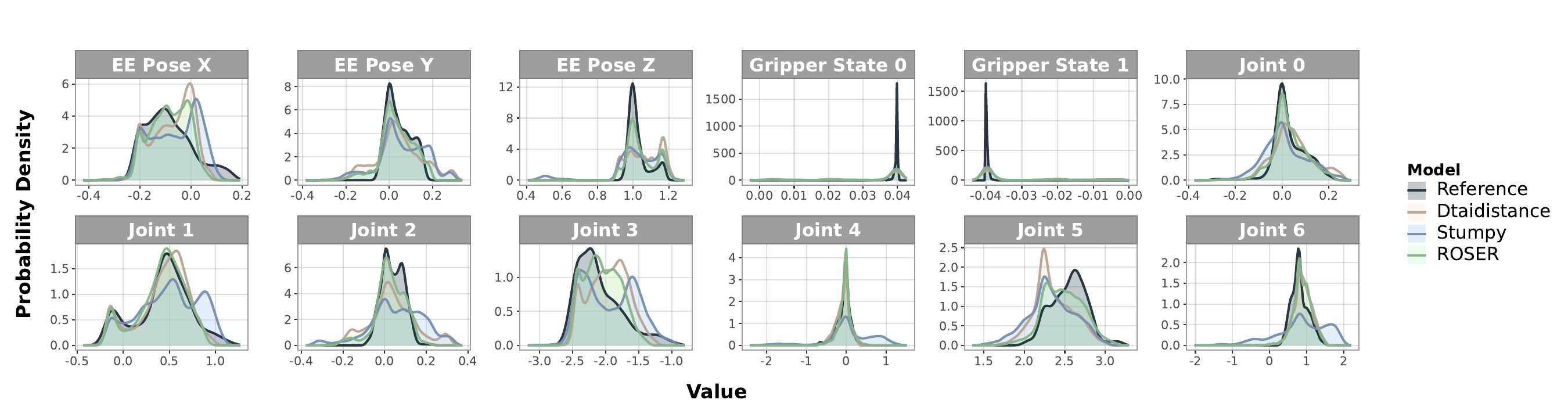}
    \caption{Feature-level distribution visualization for microwave open task in the LIBERO benchmark.}
    \label{fig:dist-libero-microwave_open}
\end{figure}

\begin{figure}[!ht]
    \centering
    \includegraphics[width=\linewidth]{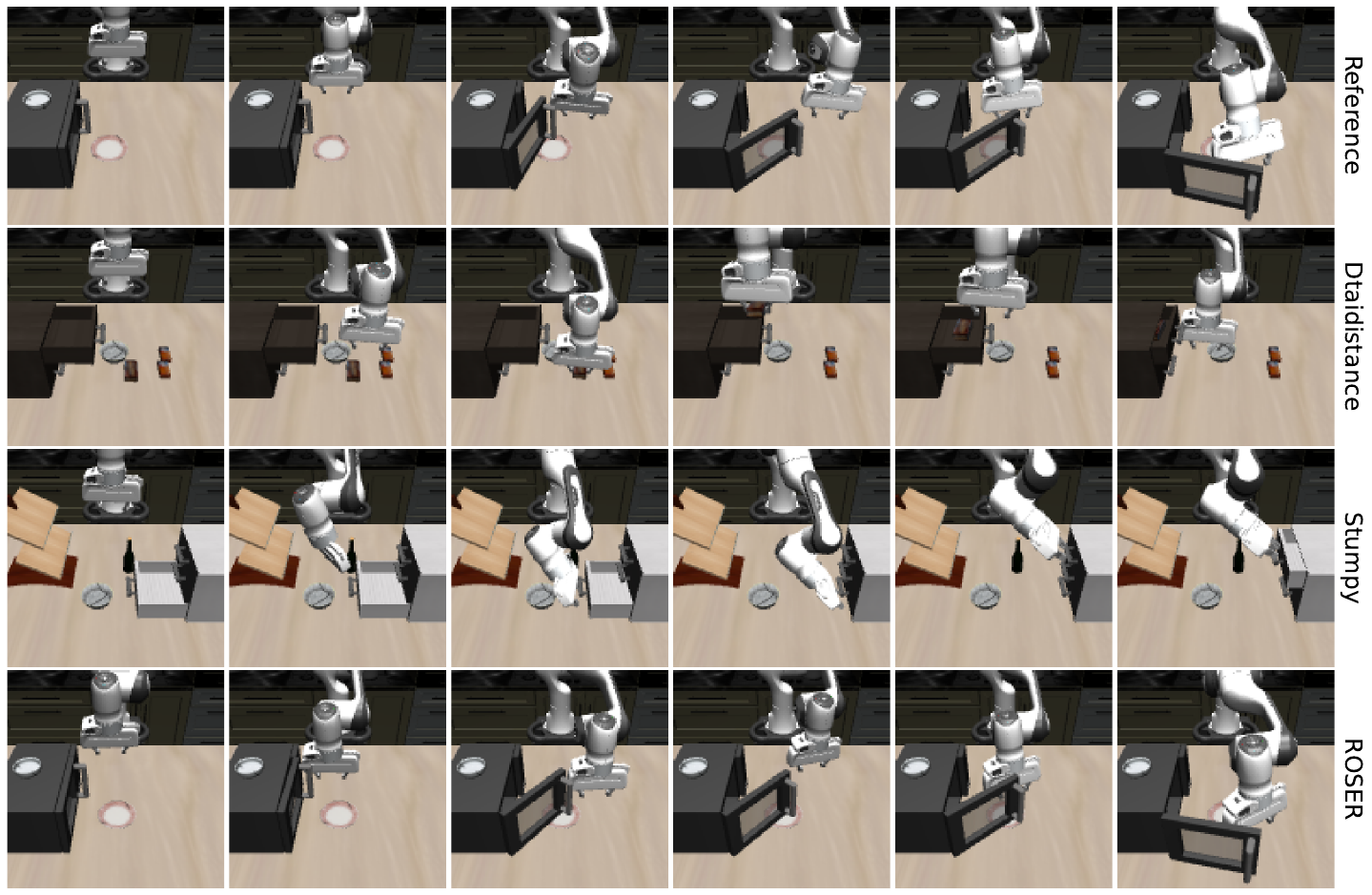}
    \caption{LIBERO qualitative results for microwave open task. ROSER retrieves correct sample while Dtaidistance and Stumpy retrieves drawer close task}
    \label{fig:quality-libero-microwave-open}
\end{figure}
\clearpage

% --- Microwave Close ---
\subsubsection{Microwave Close}
\label{sec:libero-microwave-close}

\begin{figure}[!ht]
    \centering
    \includegraphics[width=\linewidth]{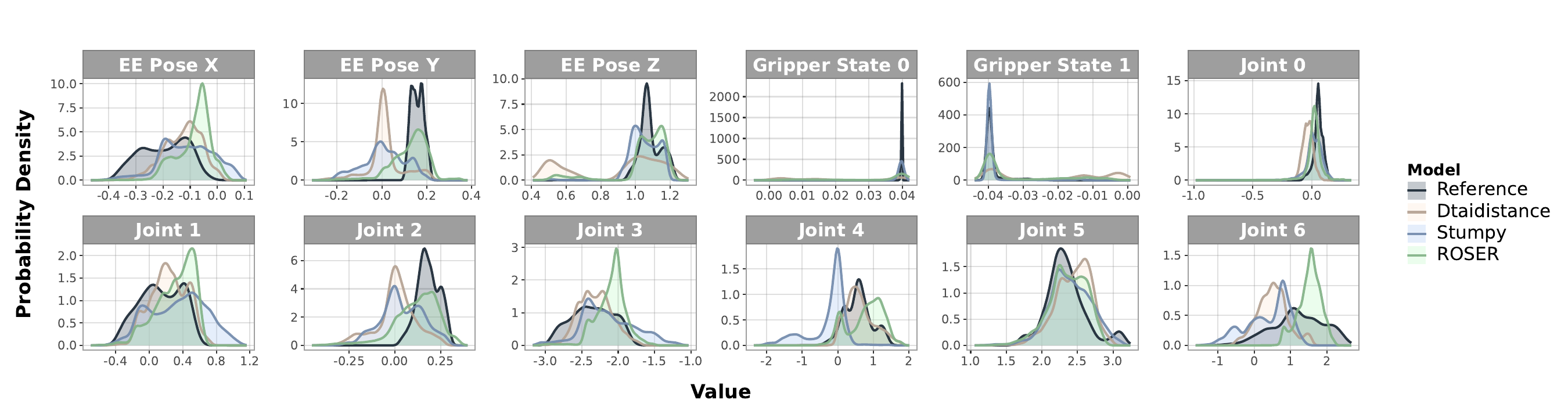}
    \caption{Feature-level distribution visualization for microwave close task in the LIBERO benchmark.}
    \label{fig:dist-libero-microwave_close}
\end{figure}

\begin{figure}[!ht]
    \centering
    \includegraphics[width=\linewidth]{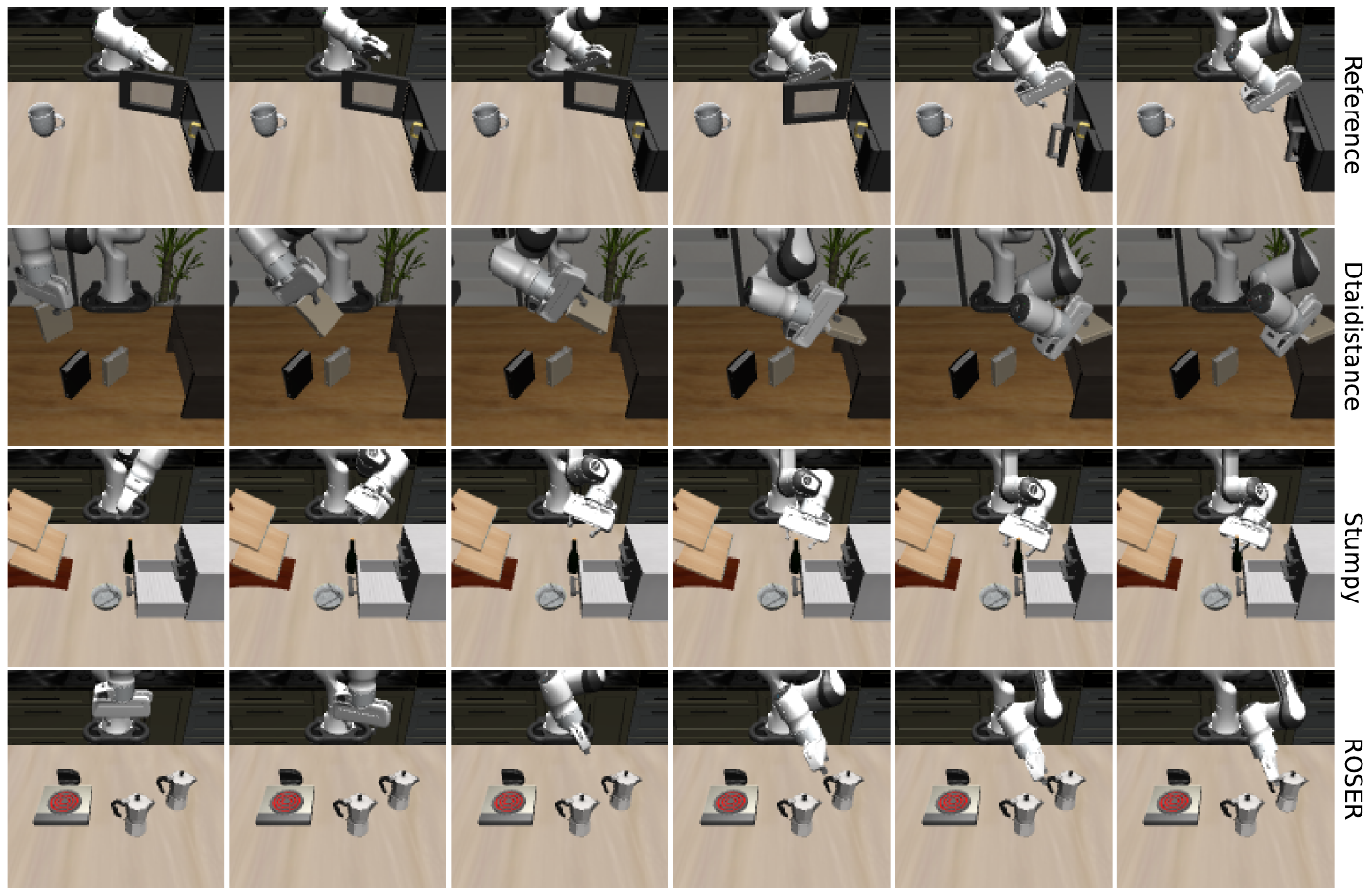}
    \caption{LIBERO qualitative results for microwave close task. All the models fail to retrieve correct tasks on this.}
    \label{fig:quality-libero-microwave-close}
\end{figure}
\clearpage

% --- Stove On ---
\subsubsection{Stove On}
\label{sec:libero-stove-on}

\begin{figure}[!ht]
    \centering
    \includegraphics[width=\linewidth]{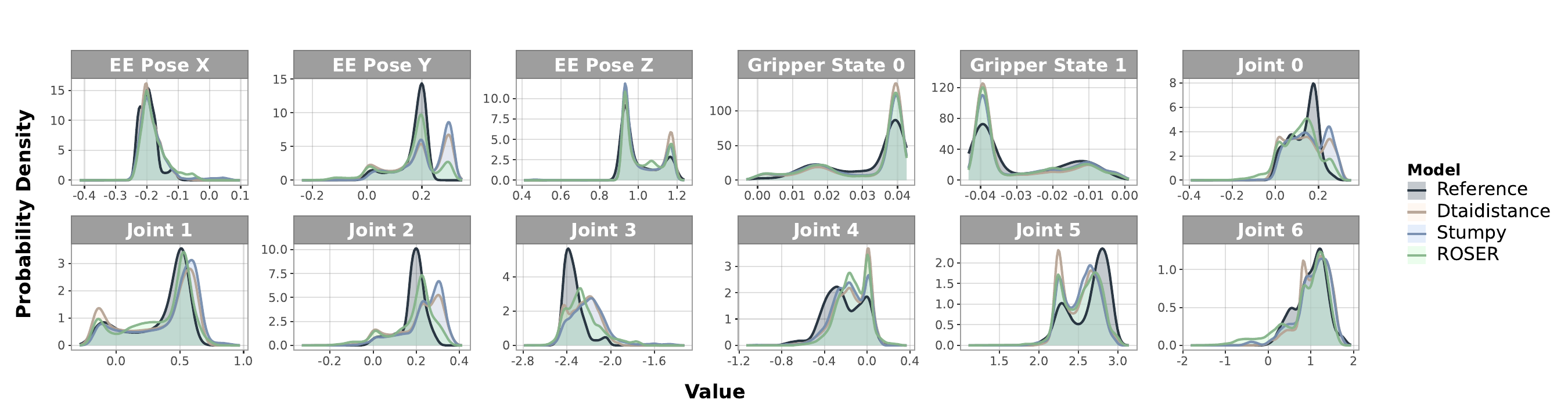}
    \caption{Feature-level distribution visualization for stove on task in the LIBERO benchmark.}
    \label{fig:dist-libero-stove_on}
\end{figure}

\begin{figure}[!ht]
    \centering
    \includegraphics[width=\linewidth]{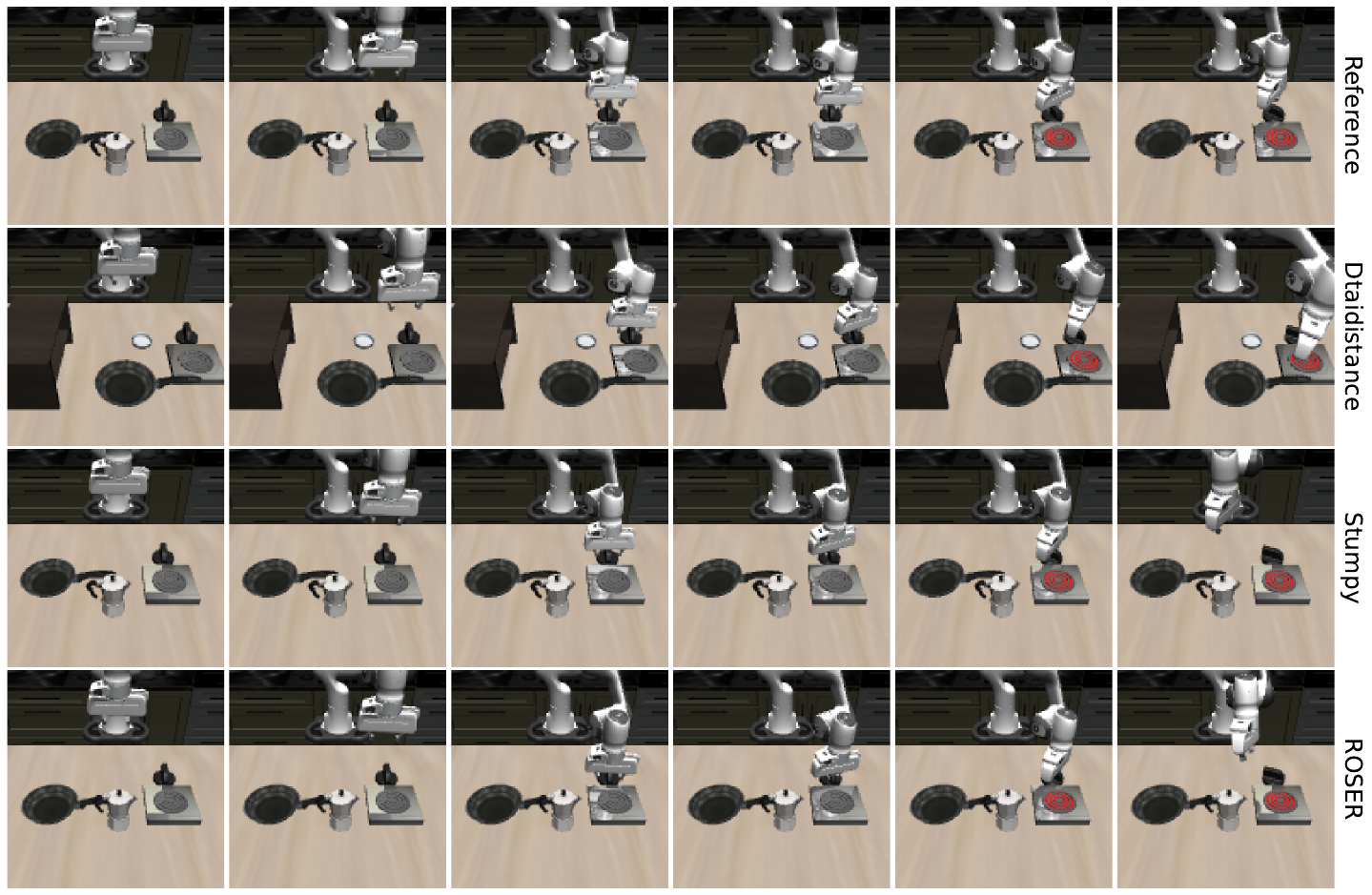}
    \caption{LIBERO qualitative results for stove on task. All models successfully retrieve samples for this task.}
    \label{fig:quality-libero-stove-on}
\end{figure}
\clearpage

% --- Stove Off ---
\subsubsection{Stove Off}
\label{sec:libero-stove-off}

\begin{figure}[!ht]
    \centering
    \includegraphics[width=\linewidth]{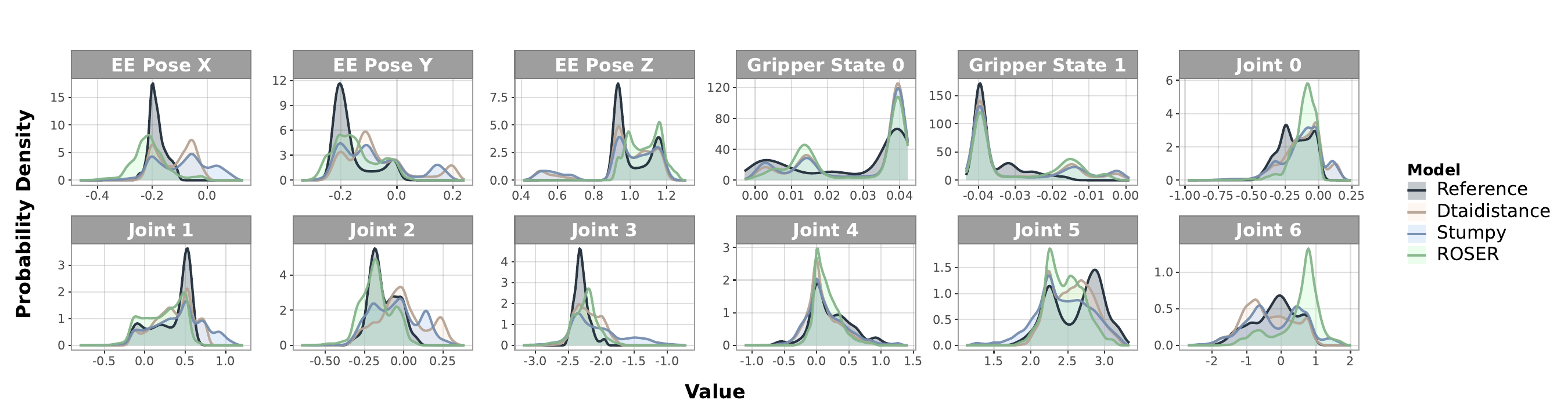}
    \caption{Feature-level distribution visualization for stove off task in the LIBERO benchmark.}
    \label{fig:dist-libero-stove_off}
\end{figure}

\begin{figure}[!ht]
    \centering
    \includegraphics[width=\linewidth]{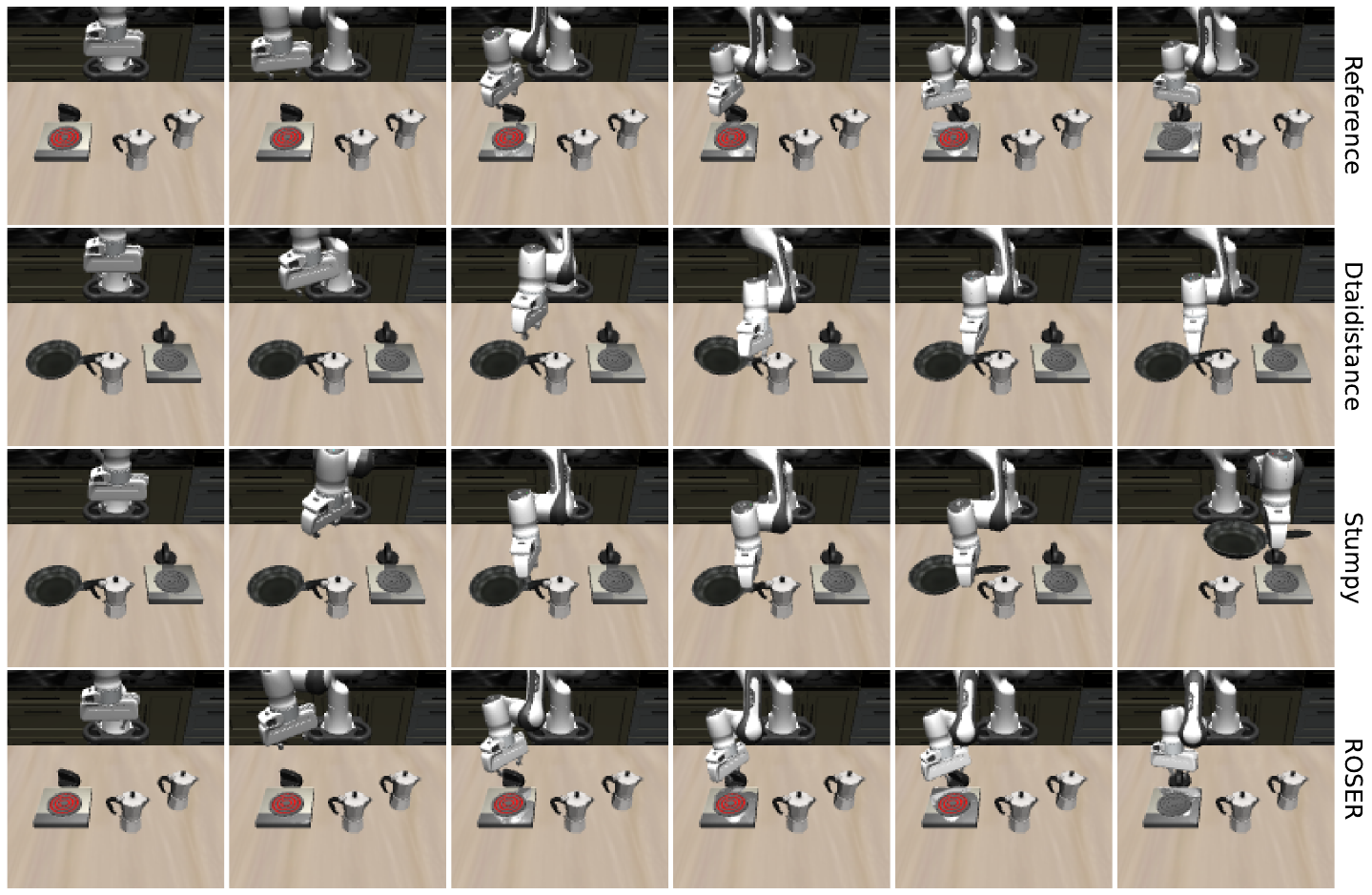}
    \caption{LIBERO qualitative results for stove off. ROSER successfully retrieves stove off tasks while Dtaidistance and Stumpy retrieves pick and place tasks}
    \label{fig:quality-libero-stove-off}
\end{figure}
\clearpage

% --- Top Drawer Open ---
\subsubsection{Top Drawer Open}
\label{sec:libero-top-drawer-open}

\begin{figure}[!ht]
    \centering
    \includegraphics[width=\linewidth]{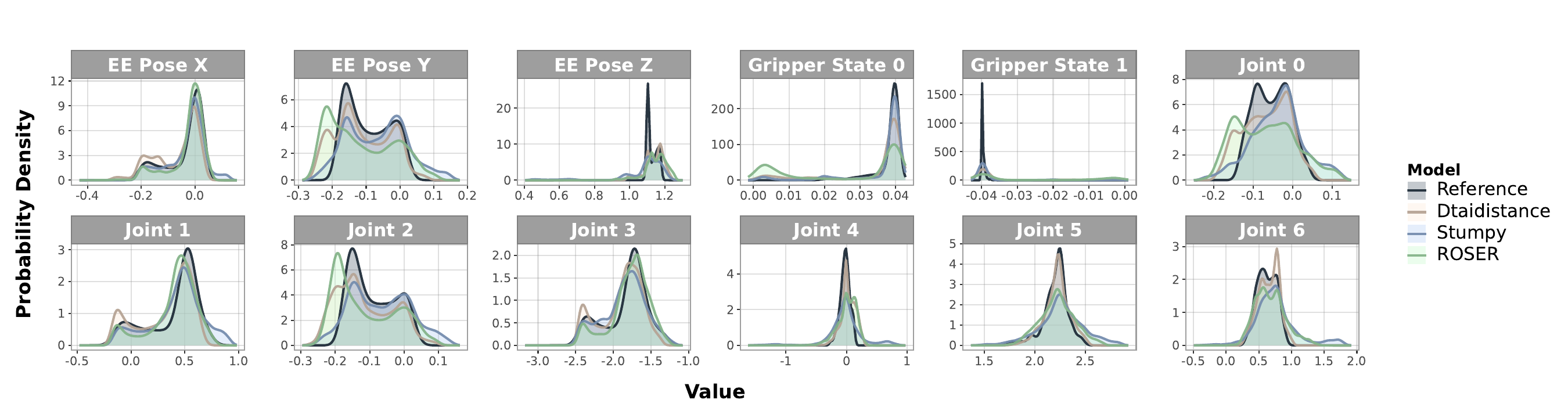}
    \caption{Feature-level distribution visualization for top drawer open task in the LIBERO benchmark.}
    \label{fig:dist-libero-top_drawer_open}
\end{figure}

\begin{figure}[!ht]
    \centering
    \includegraphics[width=\linewidth]{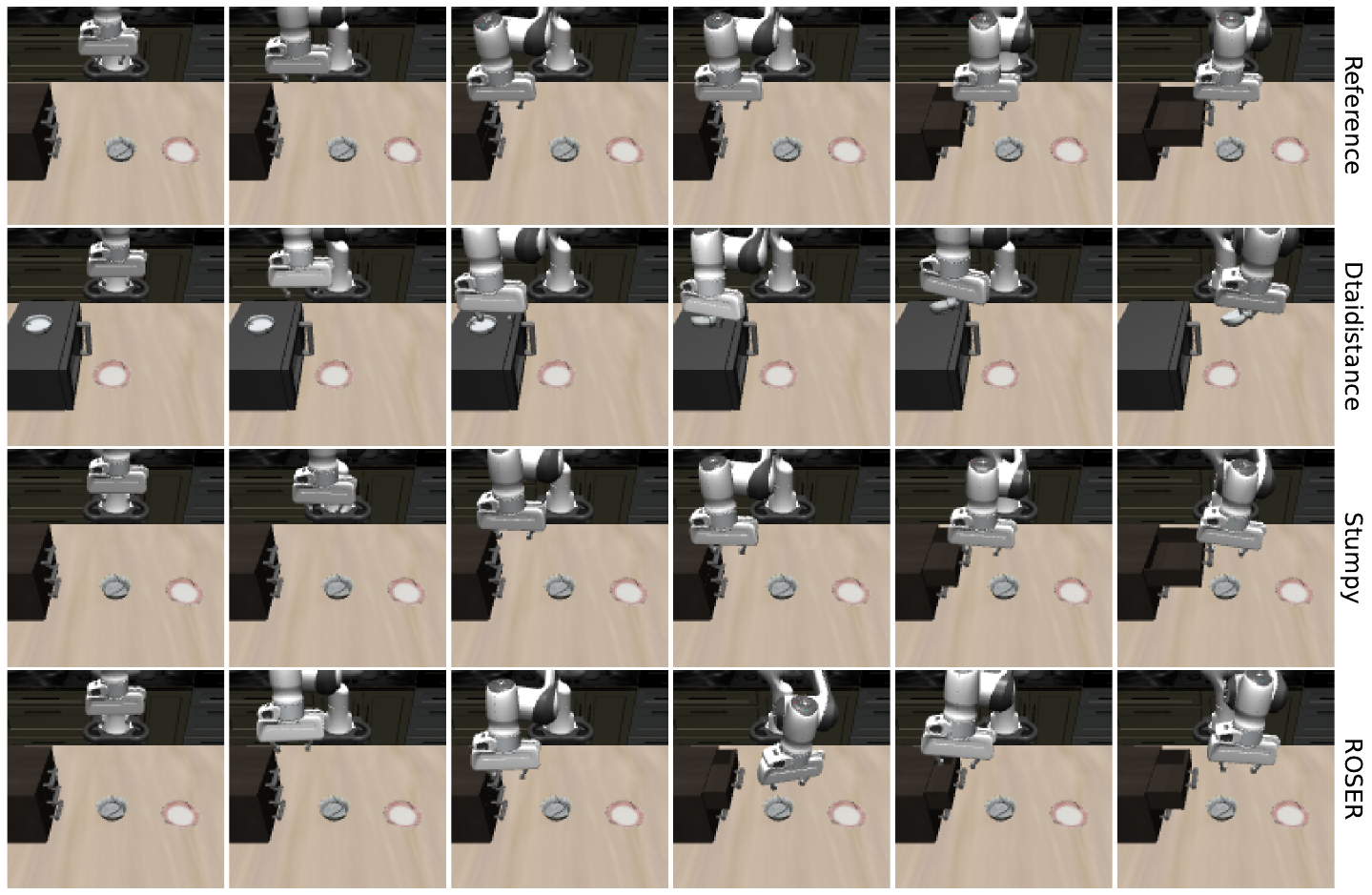}
    \caption{LIBERO qualitative results for top drawer open. ROSER and Stumpy successfully retrieve the task while Dtaidistance retrieves pick and place task.}
    \label{fig:quality-libero-top-drawer-open}
\end{figure}
\clearpage

% --- Top Drawer Close ---
\subsubsection{Top Drawer Close}
\label{sec:libero-top-drawer-close}

\begin{figure}[!ht]
    \centering
    \includegraphics[width=\linewidth]{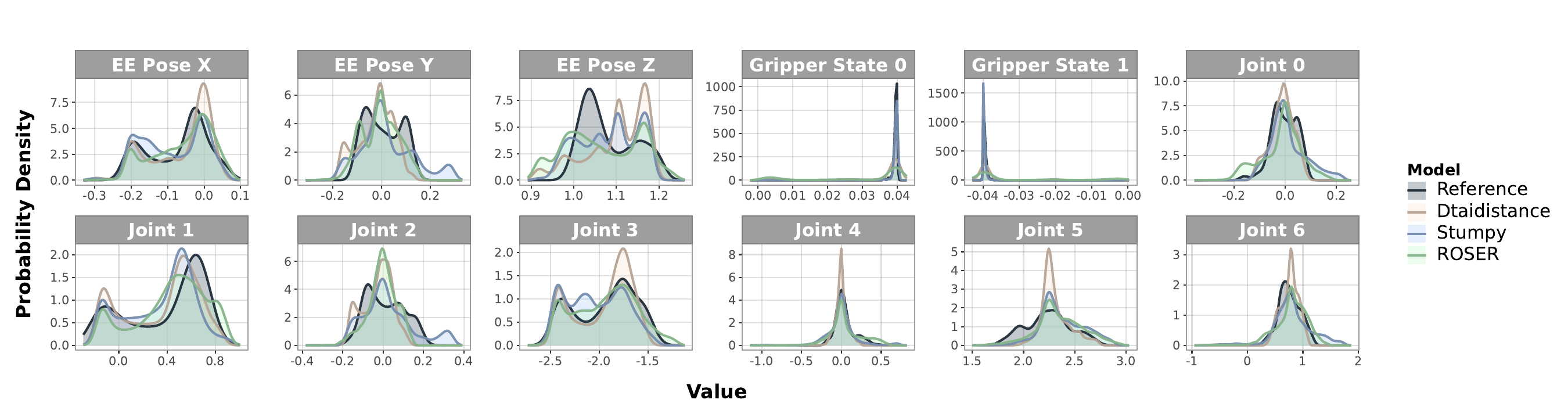}
    \caption{Feature-level distribution visualization for top drawer close task in the LIBERO benchmark.}
    \label{fig:dist-libero-top_drawer_close}
\end{figure}

\begin{figure}[!ht]
    \centering
    \includegraphics[width=\linewidth]{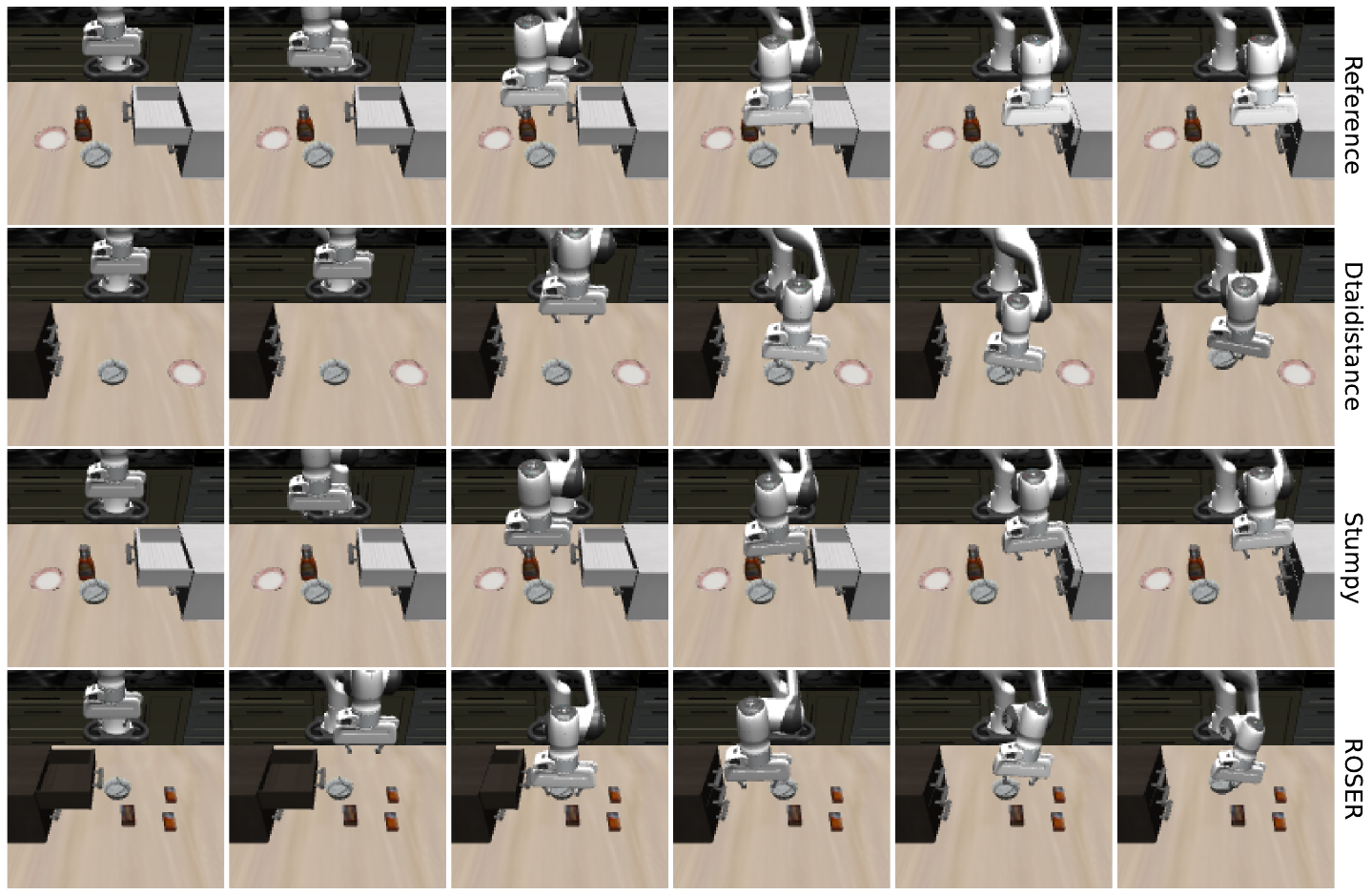}
    \caption{Libero qualitative results for top drawer close. ROSER and Stumpy successfully retrieve the task while Dtaidistance retrieves pick and place task.}
    \label{fig:quality-libero-top-drawer-close}
\end{figure}
\clearpage

% --- Bottom Drawer Open ---
\subsubsection{Bottom Drawer Open}
\label{sec:libero-bottom-drawer-open}

\begin{figure}[!ht]
    \centering
    \includegraphics[width=\linewidth]{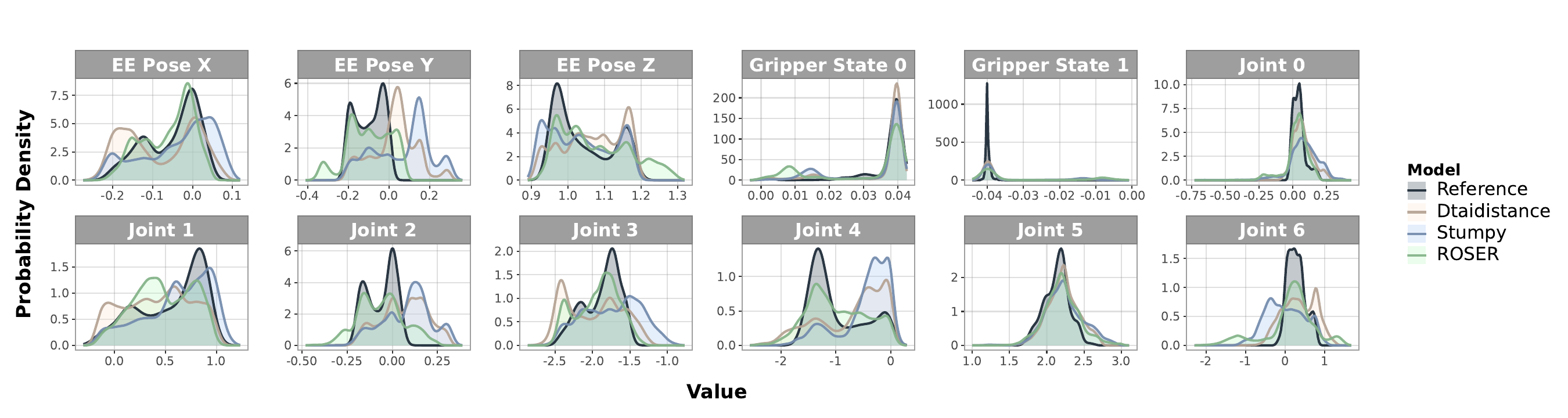}
    \caption{Feature-level distribution visualization for bottom drawer open task in the LIBERO benchmark.}
    \label{fig:dist-libero-bottom-drawer_open}
\end{figure}

\begin{figure}[!ht]
    \centering
    \includegraphics[width=\linewidth]{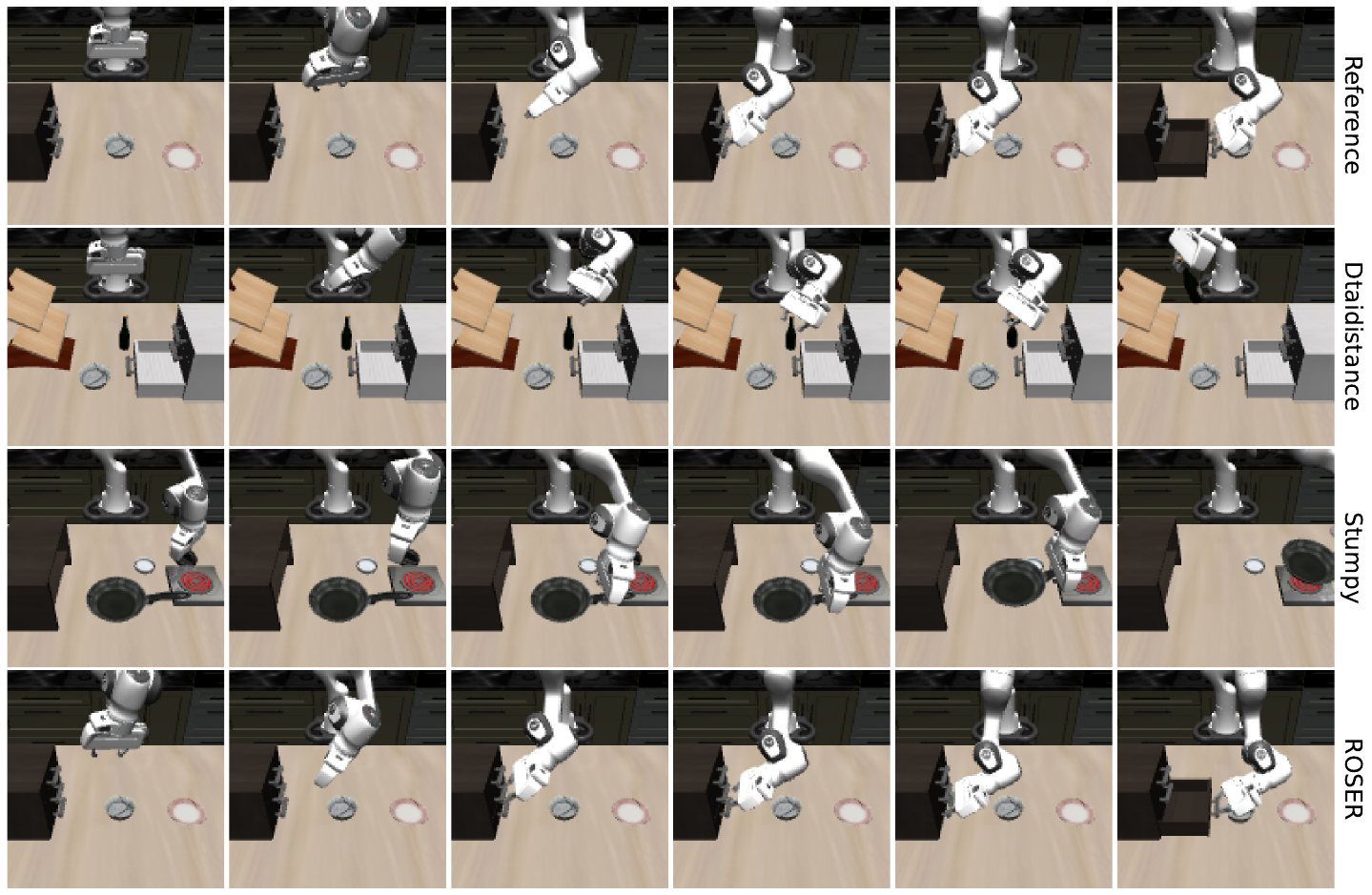}
    \caption{LIBERO qualitative results for bottom drawer open. This task is challenging since robot arms need to avoid collding with top drawer handle to reach bottom drawer handle. ROSER successfully retrieves similar task while Stumpy and Dtaidistance retrieve other tasks.}
    \label{fig:quality-libero-bottom-drawer-open}
\end{figure}
\clearpage

% --- Bottom Drawer Close ---
\subsubsection{Bottom Drawer Close}
\label{sec:libero-bottom-drawer-close}

\begin{figure}[!ht]
    \centering
    \includegraphics[width=\linewidth]{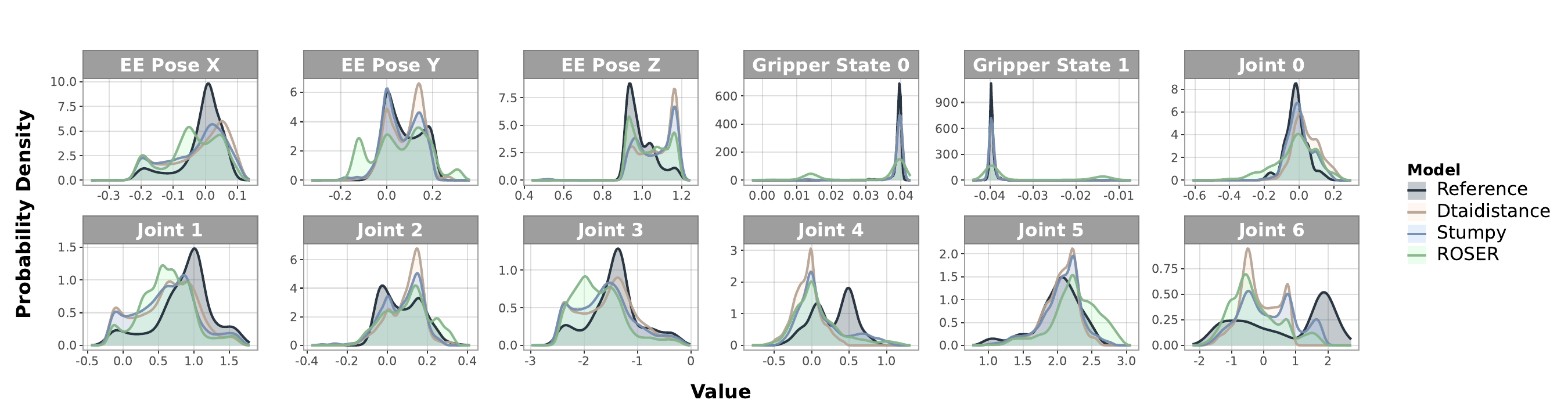}
    \caption{Feature-level distribution visualization for bottom drawer close task in the LIBERO benchmark.}
    \label{fig:dist-libero-bottom_drawer_close}
\end{figure}

\begin{figure}[!ht]
    \centering
    \includegraphics[width=\linewidth]{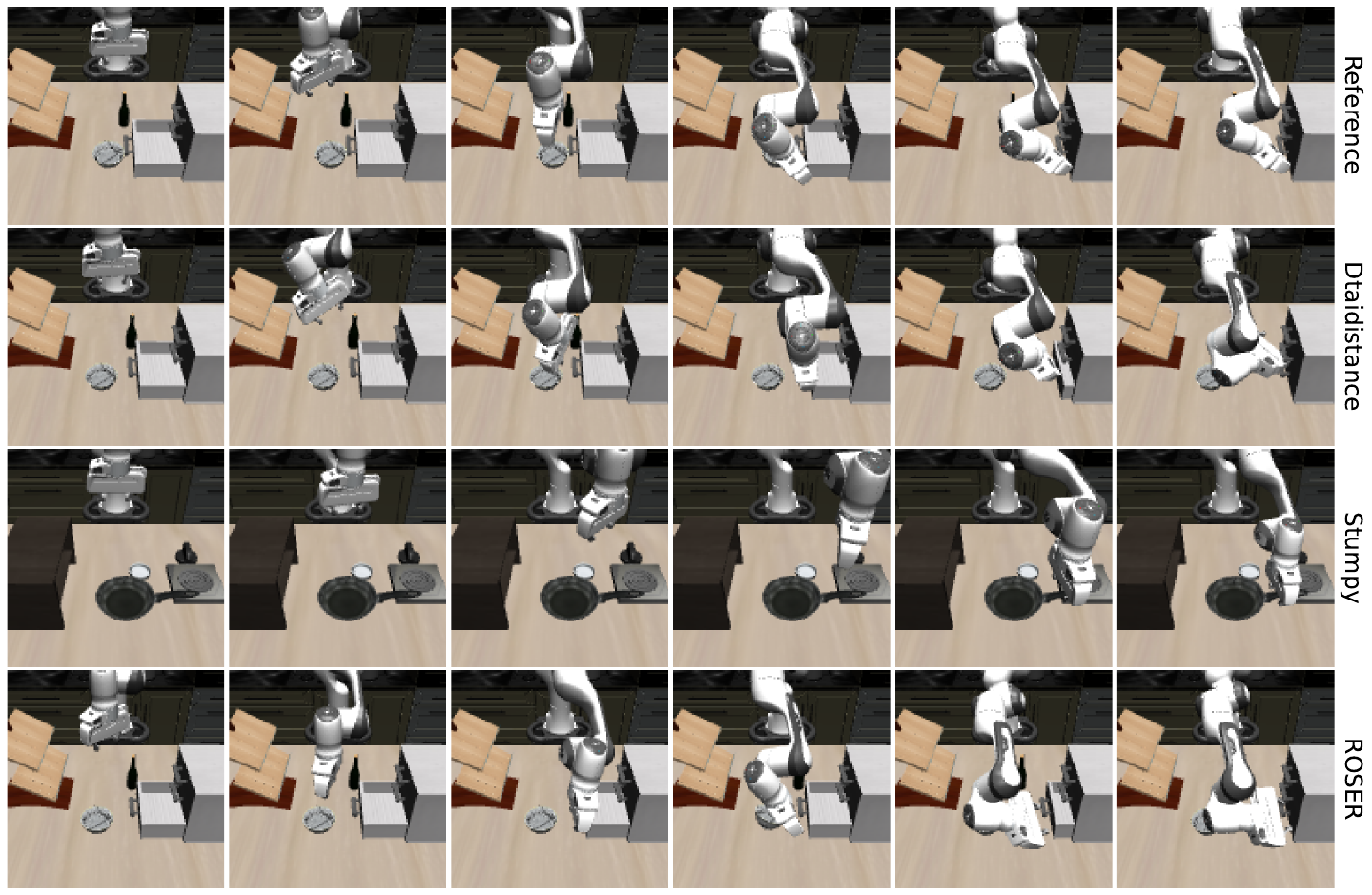}
    \caption{LIBERO qualitative results for bottom drawer close. ROSER and Dtaidistance successfully retrieves similar task while Stumpy retrieves pick and place task.}
    \label{fig:quality-libero-bottom-drawer-close}
\end{figure}
\clearpage

% --- Pick and Place (PnP) ---
\subsubsection{Pick and Place (PnP)}
\label{sec:libero-pnp}

\begin{figure}[!ht]
    \centering
    \includegraphics[width=\linewidth]{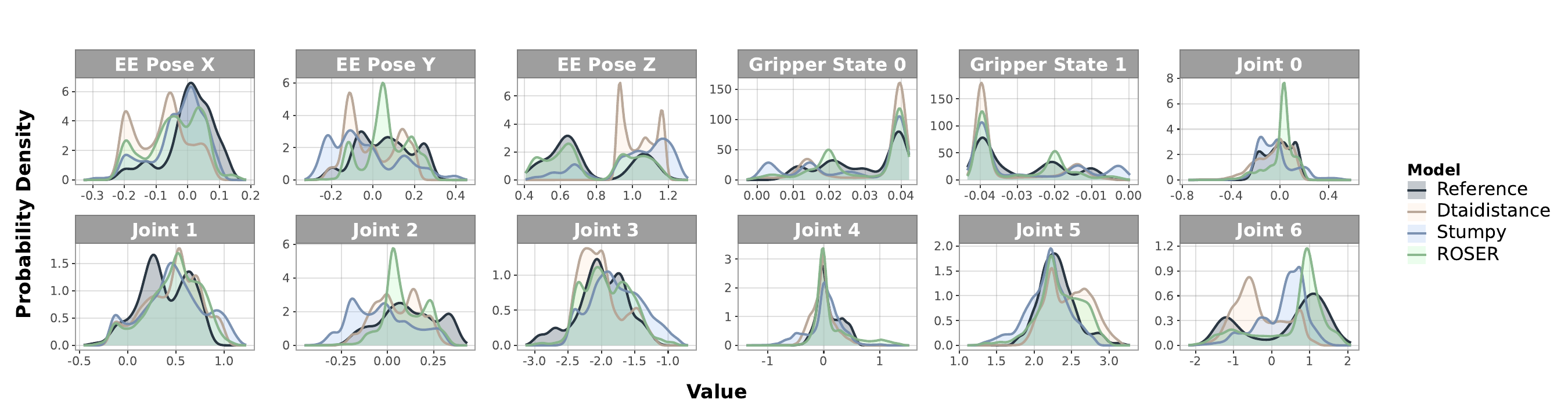}
    \caption{Feature-level distribution visualization for pick and place task in the LIBERO benchmark.}
    \label{fig:dist-libero-pnp}
\end{figure}

\begin{figure}[!ht]
    \centering
    \includegraphics[width=\linewidth]{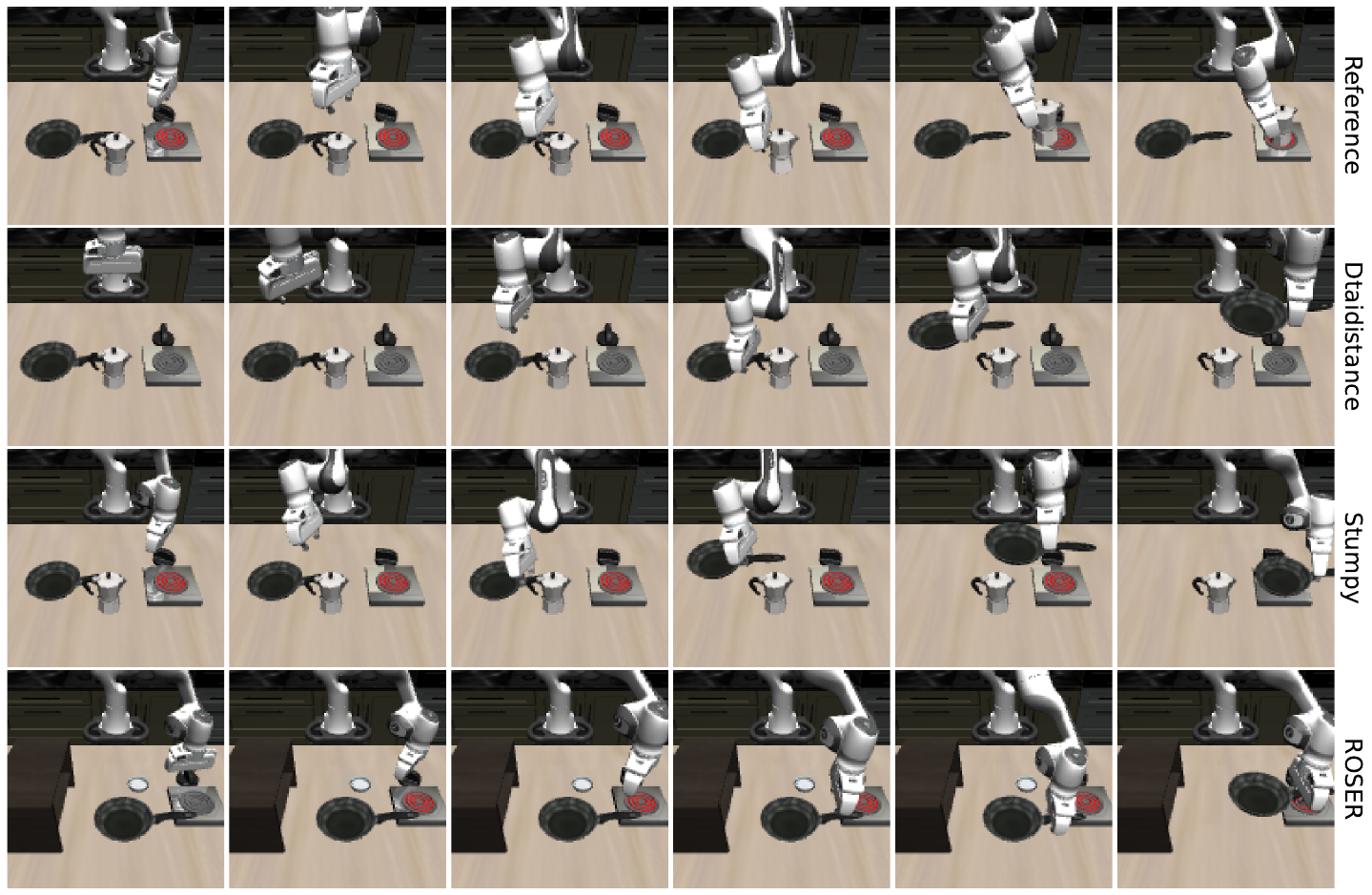}
    \caption{LIBERO qualitative results pick and place. All models can successfully retrieve pick an place tasks}
    \label{fig:quality-libero-pick-and-place}
\end{figure}
\clearpage

\subsection{DROID Benchmark}
\label{sec:appendix-droid}

% --- Close Drawer ---
\subsubsection{Close Drawer}
\begin{figure}[!ht]
    \centering
    \includegraphics[width=\linewidth]{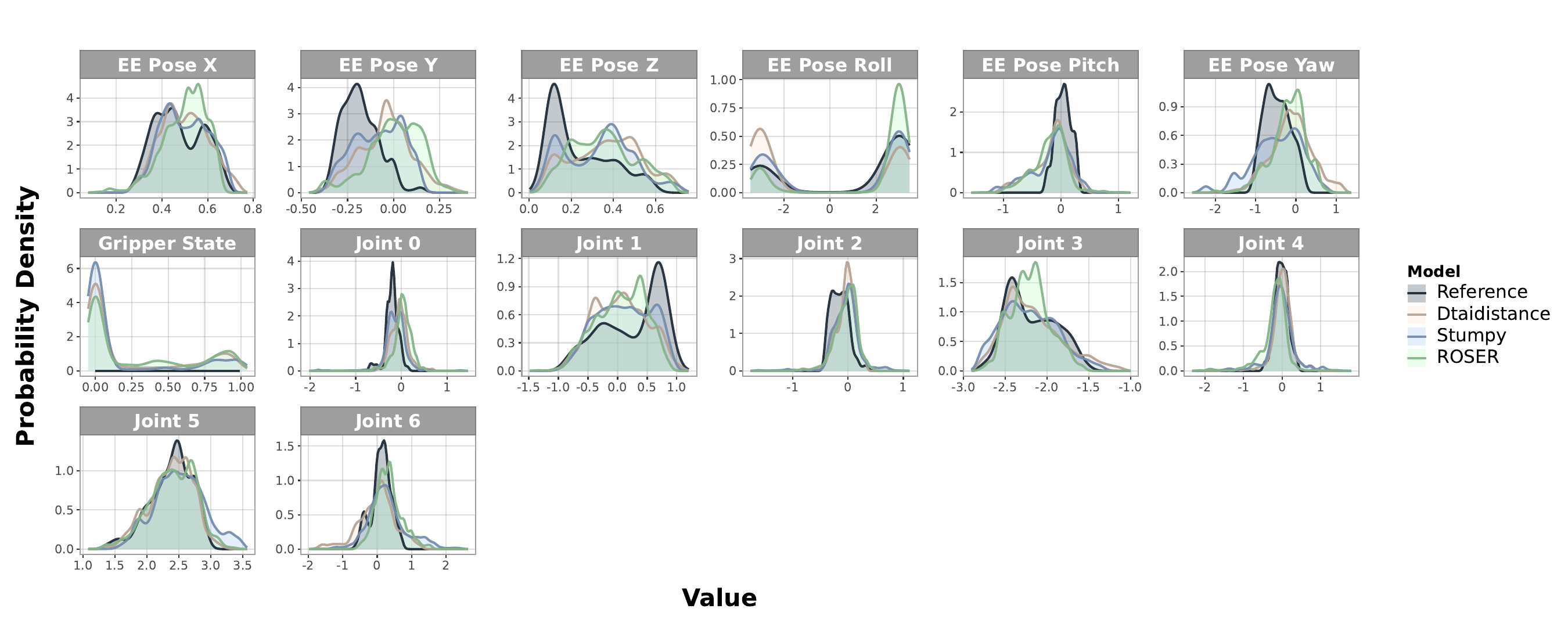}
    \caption{Feature-level distribution visualization for drawer close task in the DROID benchmark.}
    \label{fig:dist-droid-close_drawer}
\end{figure}

\begin{figure}[!ht]
    \centering
    \includegraphics[width=\linewidth]{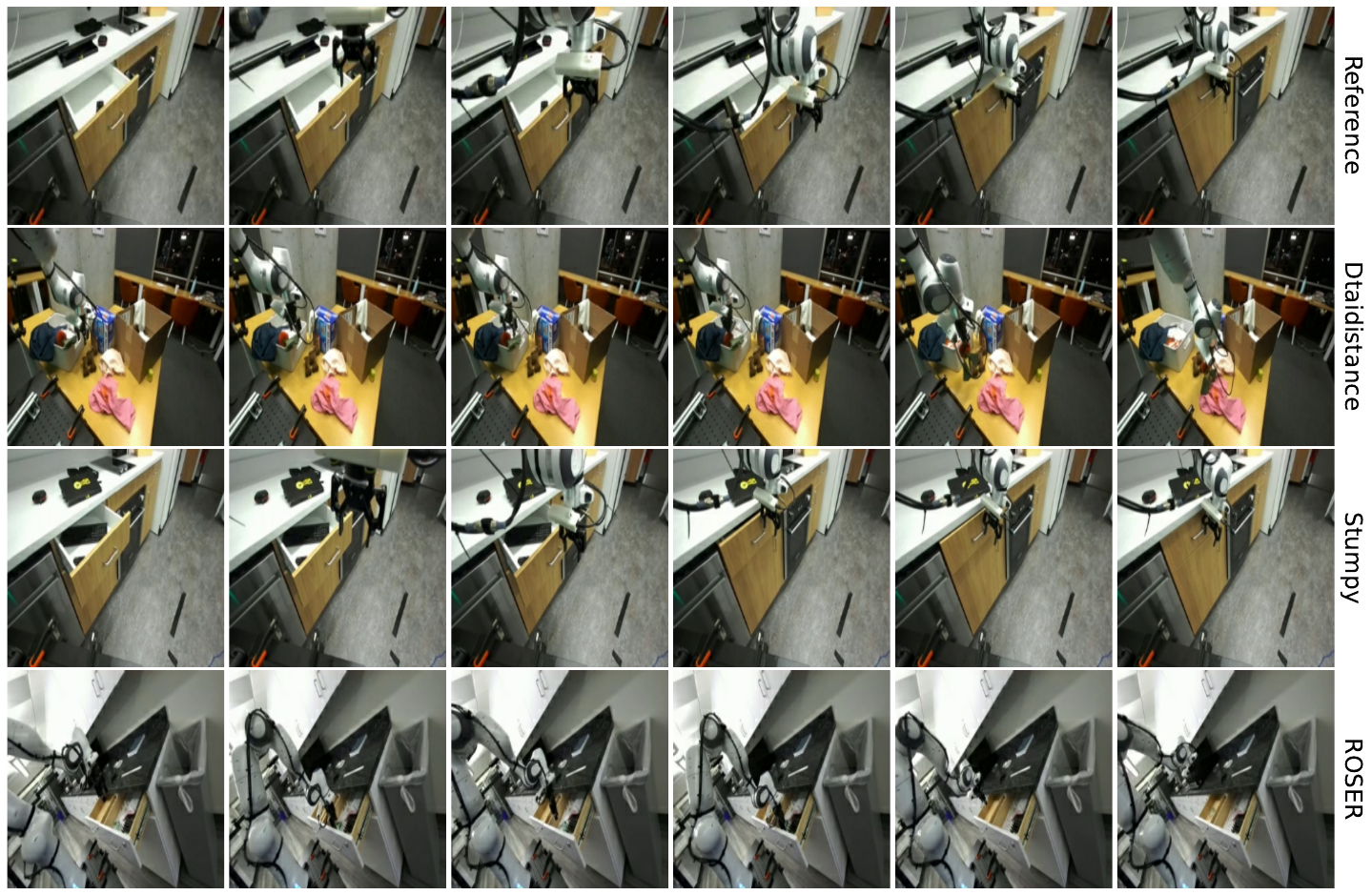}
    \caption{DROID qualitative results for drawer close. ROSER can successfully retrieve similar task even when target environment is different. }
    \label{fig:quality-droid-close-drawer}
\end{figure}
\clearpage

% --- Pick and Place (PnP) ---
\subsubsection{Pick and Place (PnP)}
\begin{figure}[!ht]
    \centering
    \includegraphics[width=\linewidth]{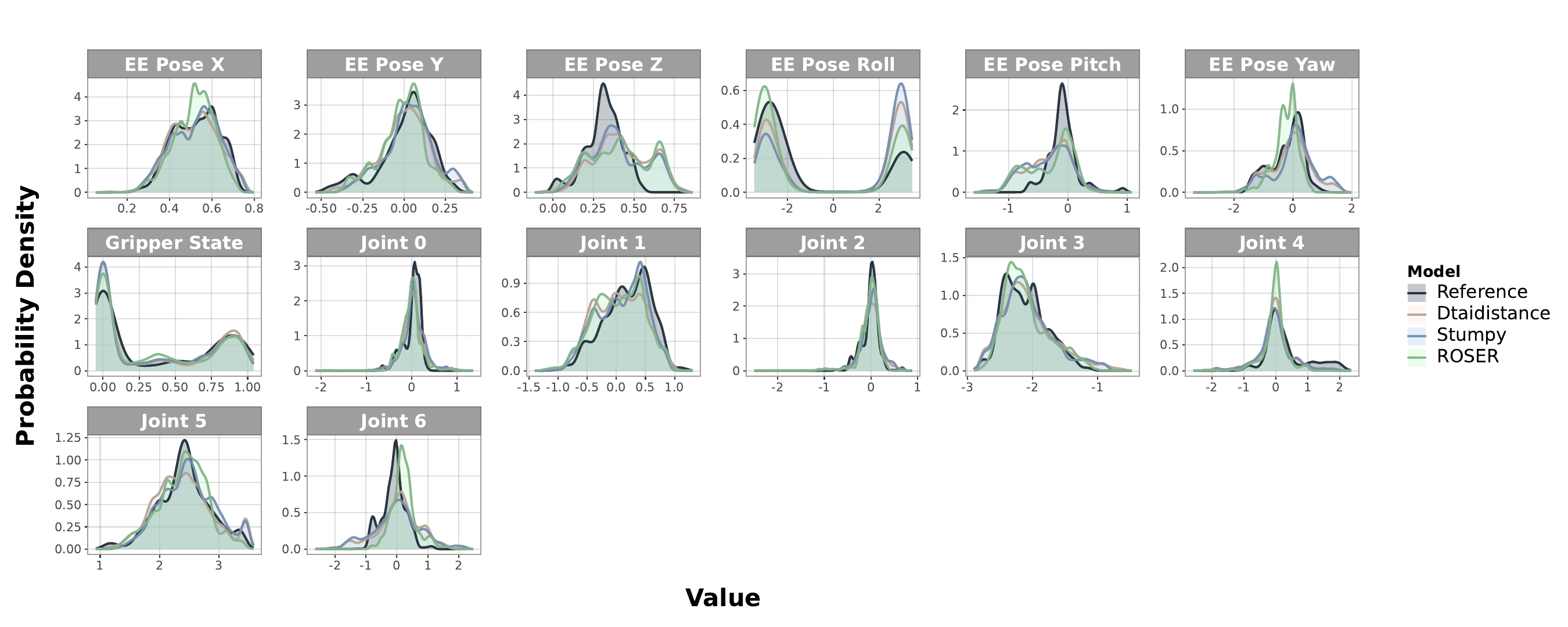}
    \caption{Feature-level distribution visualization for pick and place task in the DROID benchmark.}
    \label{fig:dist-droid-pnp}
\end{figure}

\begin{figure}[!ht]
    \centering
    \includegraphics[width=\linewidth]{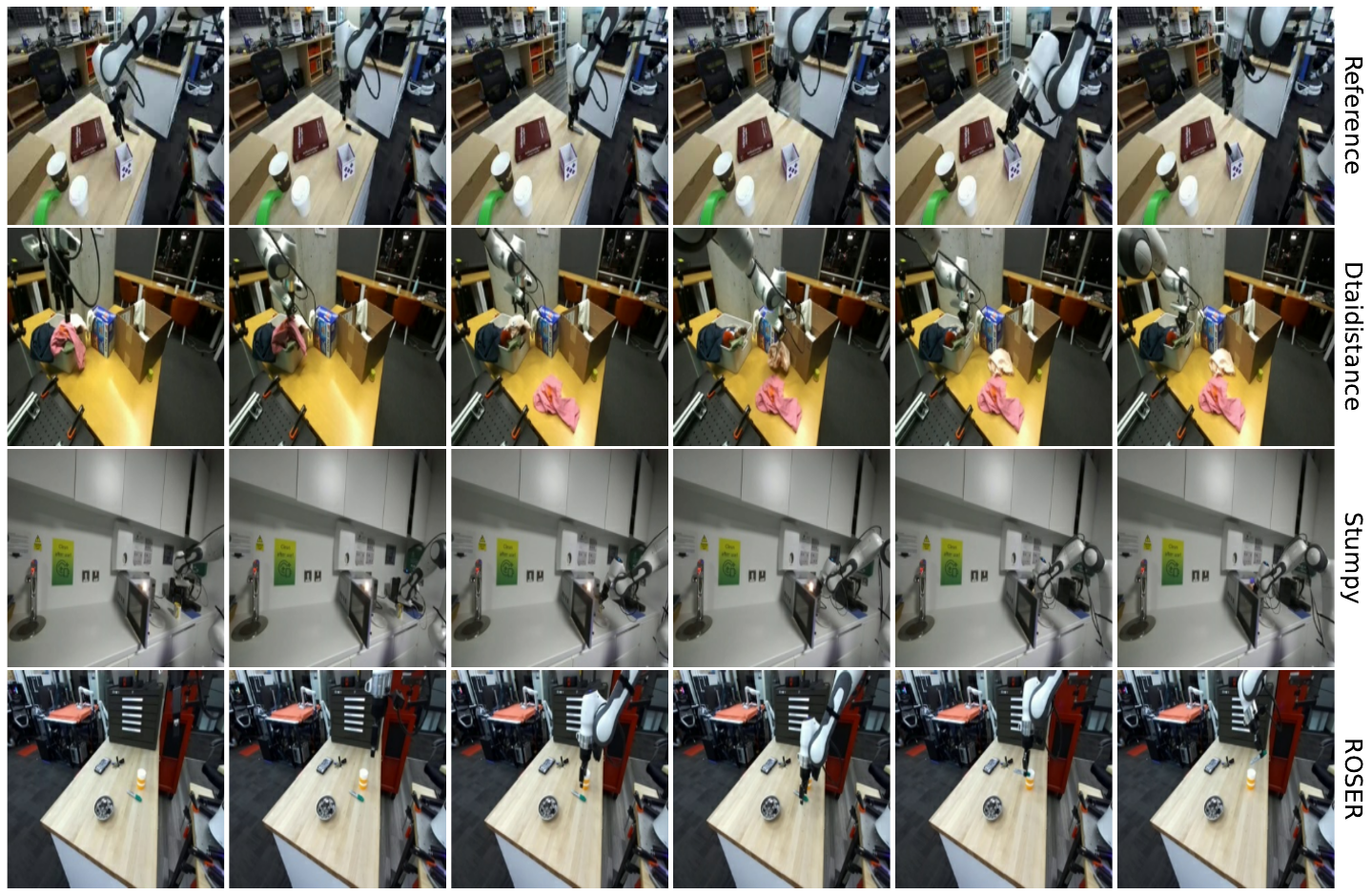}
    \caption{DROID qualitative results for pick and place task. All models successfully retrieve similar tasks.}
    \label{fig:quality-droid-pick-and-place}
\end{figure}
\clearpage

% --- Turn Faucet ---
\subsubsection{Turn Faucet}
\begin{figure}[!ht]
    \centering
    \includegraphics[width=\linewidth]{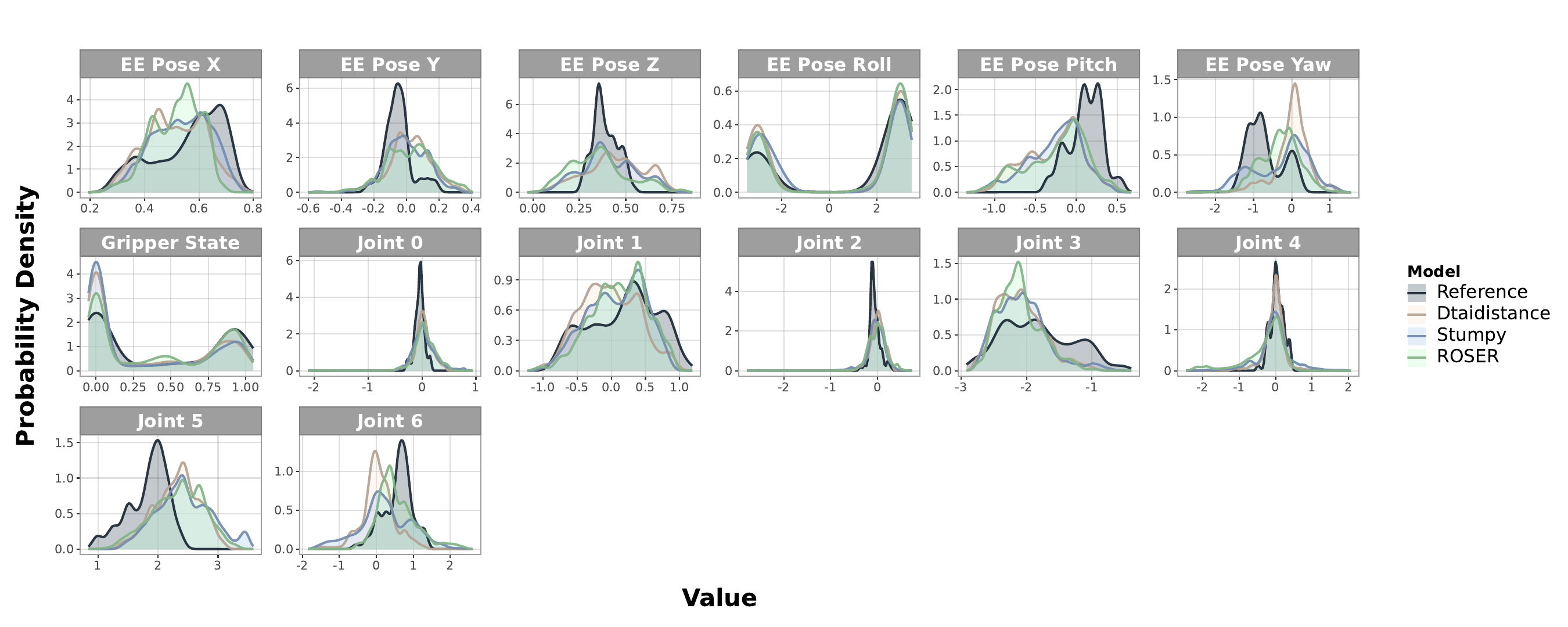}
    \caption{Feature-level distribution visualization for turn faucet task in the DROID benchmark.}
    \label{fig:dist-droid-turn}
\end{figure}

\begin{figure}[!ht]
    \centering
    \includegraphics[width=\linewidth]{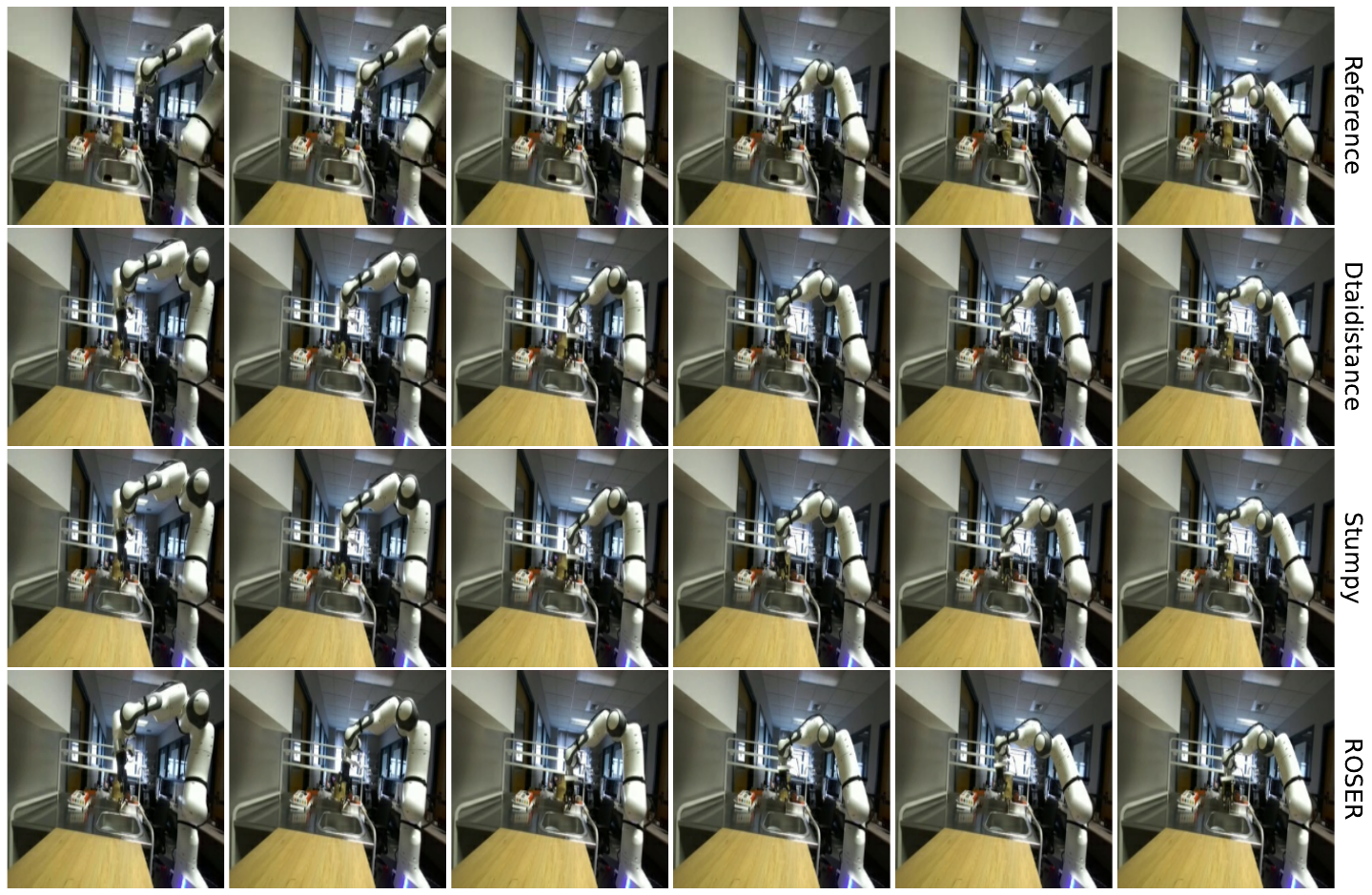}
    \caption{DROID qualitative results on turn faucet. All models successfully retrieve similar tasks.}
    \label{fig:quality-droid-turn}
\end{figure}
\clearpage

% --- Close Cabinet ---
\subsubsection{Close Cabinet}
\begin{figure}[!ht]
    \centering
    \includegraphics[width=\linewidth]{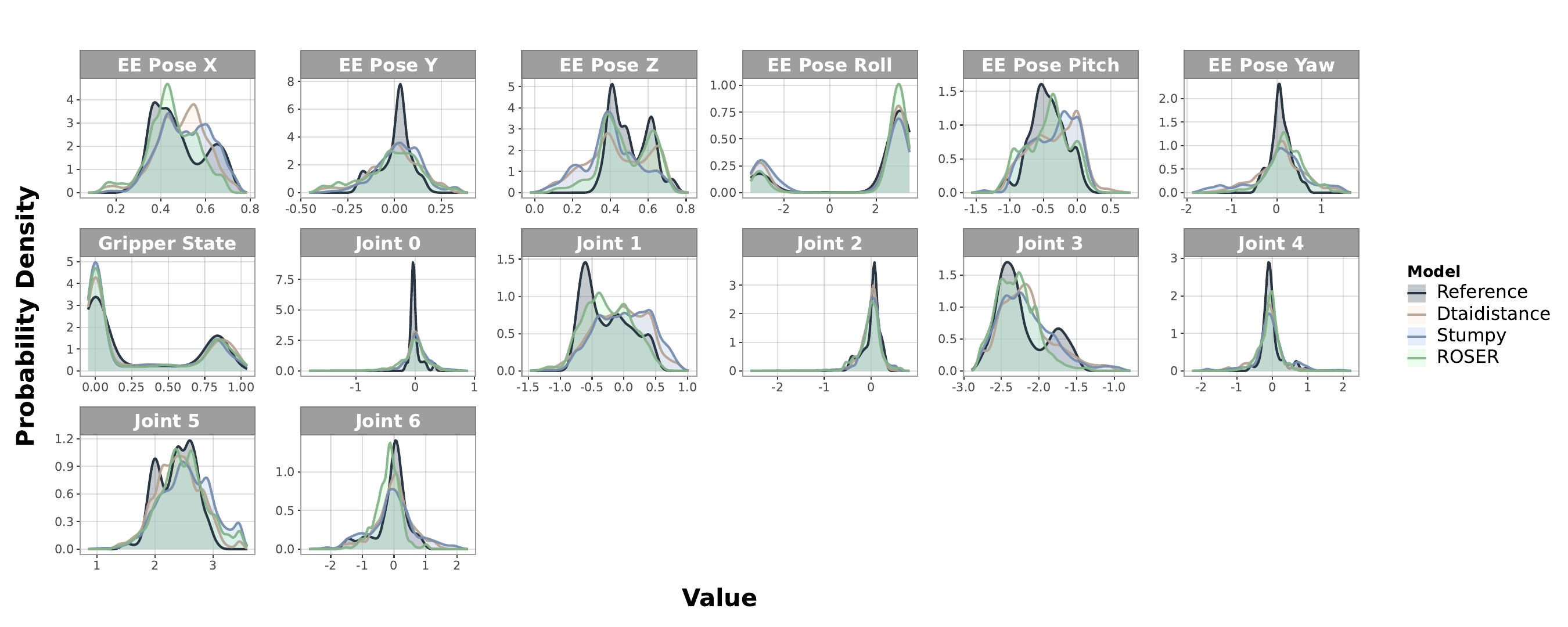}
    \caption{Feature-level distribution visualization for close cabinet task in the DROID benchmark.}
    \label{fig:dist-droid-close_cabinet}
\end{figure}

\begin{figure}[!ht]
    \centering
    \includegraphics[width=\linewidth]{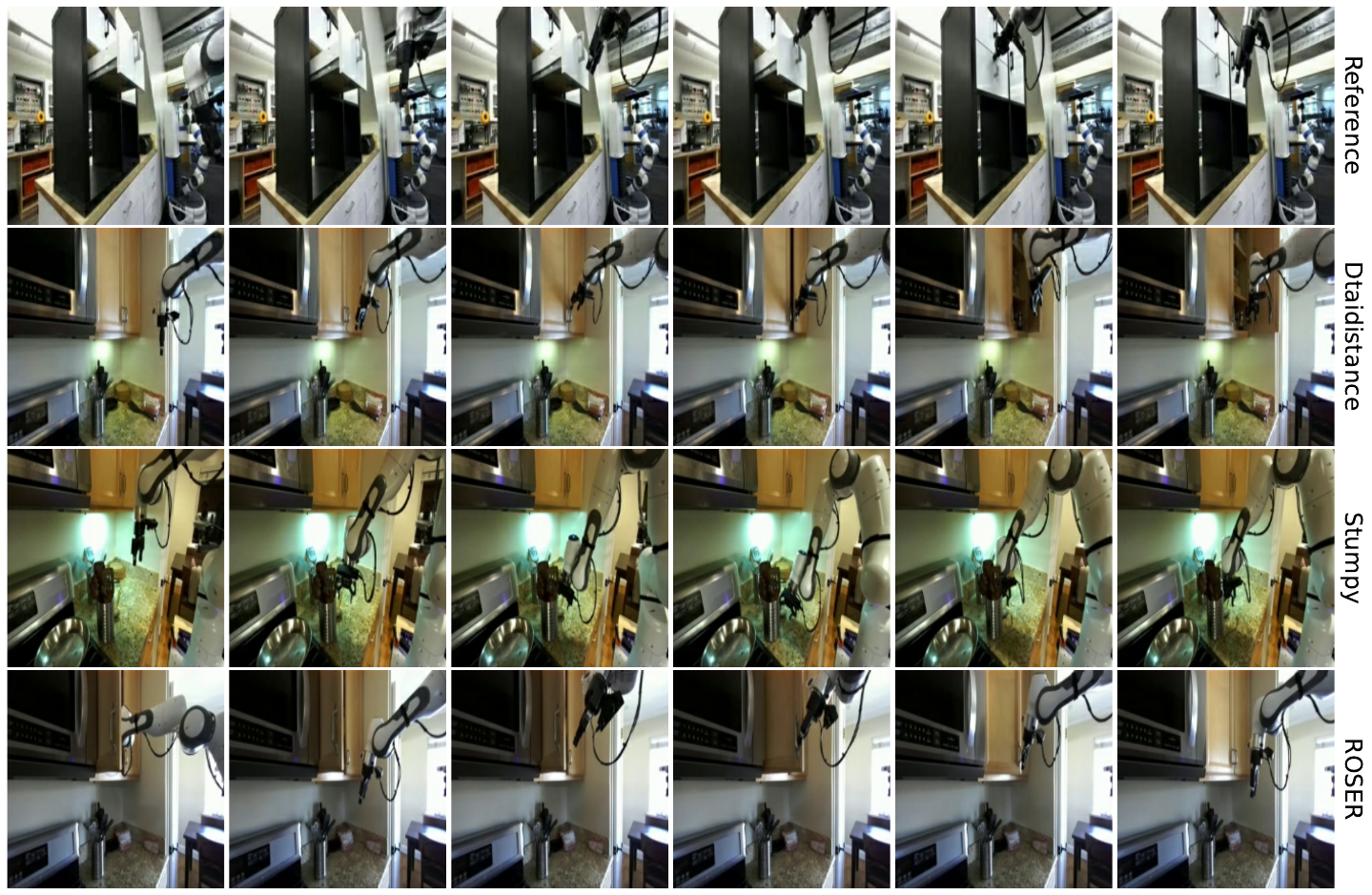}
    \caption{DROID qualitative results for close cabinet task. ROSER successfully retrieves similar task from a different environment while Stumpy retrieves a pick and place task and Dtaidistance retrieves a close cabinet task.}
    \label{fig:quality-droid-close-cabinet}
\end{figure}
\clearpage

% --- Open Cabinet ---
\subsubsection{Open Cabinet}
\begin{figure}[!ht]
    \centering
    \includegraphics[width=\linewidth]{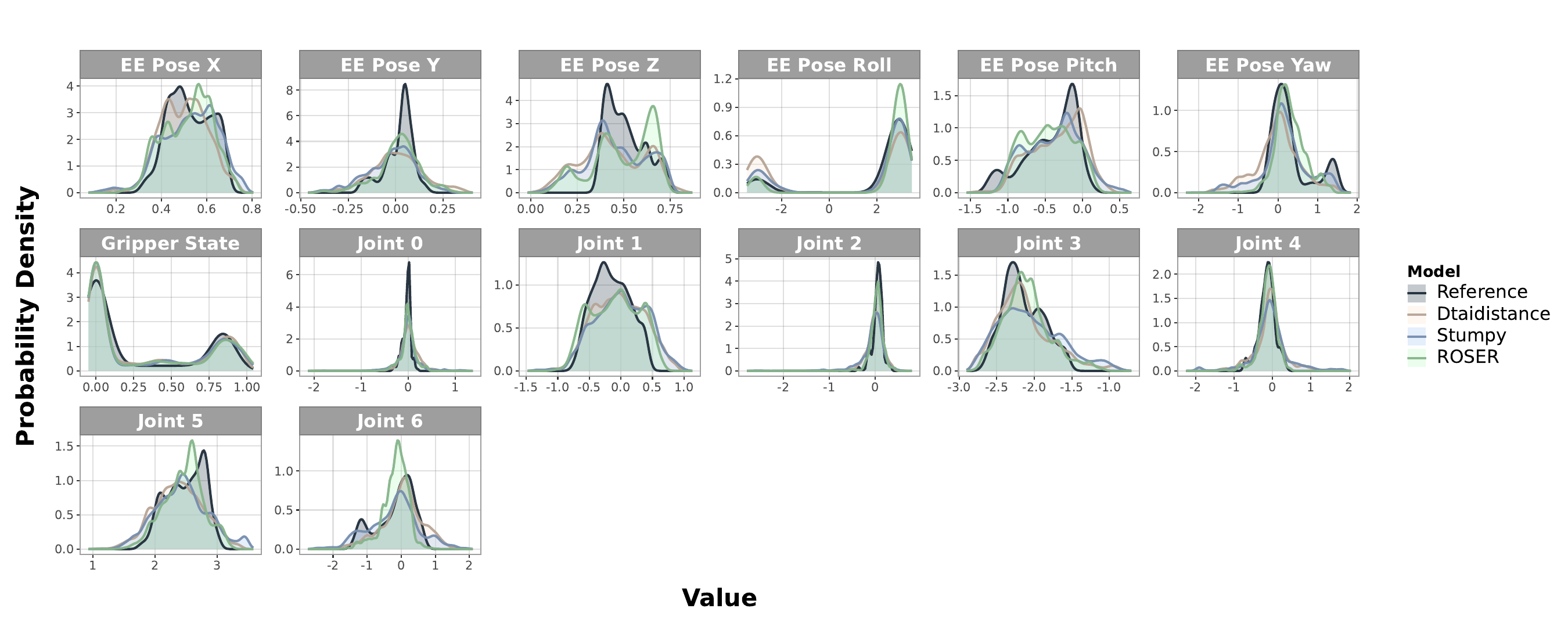}
    \caption{Feature-level distribution visualization for open cabinet task in the DROID benchmark.}
    \label{fig:dist-droid-open_cabinet}
\end{figure}

\begin{figure}[!ht]
    \centering
    \includegraphics[width=\linewidth]{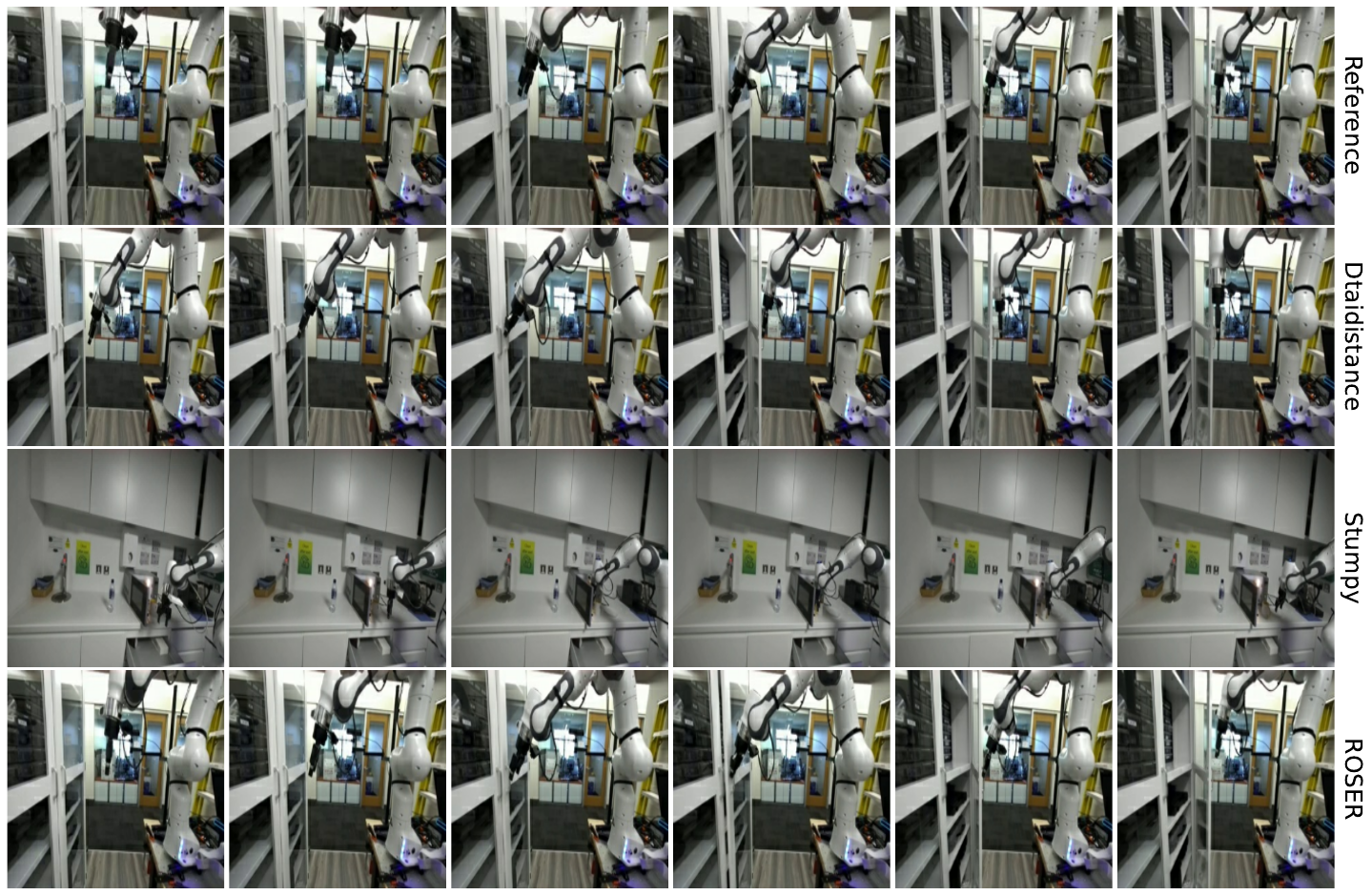}
    \caption{DROID qualitative results for open cabinet. ROSER and Dtaidistance successfully retrieve similar task while Stumpy retrieves a pick and place task.}
    \label{fig:quality-droid-open-cabinet}
\end{figure}
\clearpage

% ================= nuScenes BENCHMARK =================
\subsection{nuScenes Benchmark}
\label{sec:appendix-nuscene}

% --- Left Turn ---
\subsubsection{Left Turn}
\begin{figure}[!ht]
    \centering
    \includegraphics[width=\linewidth]{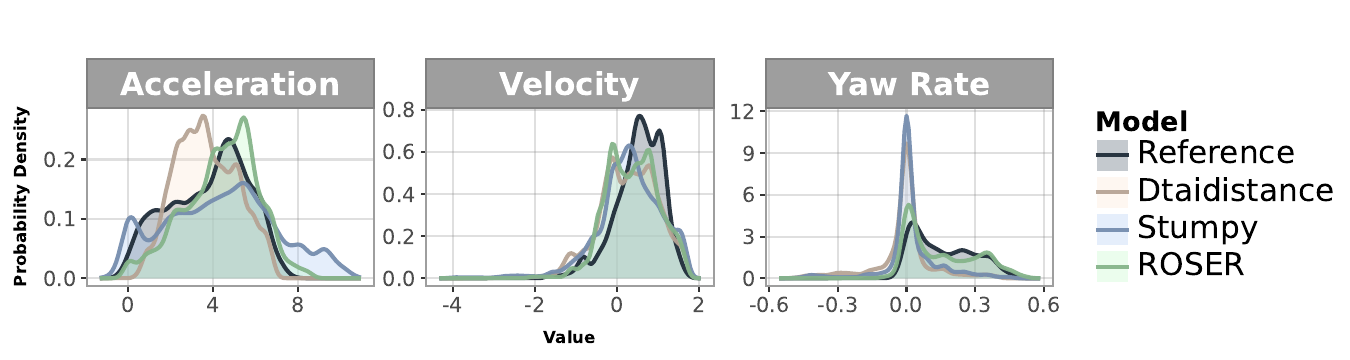}
    \caption{Feature-level distribution visualization for left turn maneuver in the nuScenes benchmark.}
    \label{fig:dist-nuscene-left_turn}
\end{figure}

\begin{figure}[!ht]
    \centering
    \includegraphics[width=\linewidth]{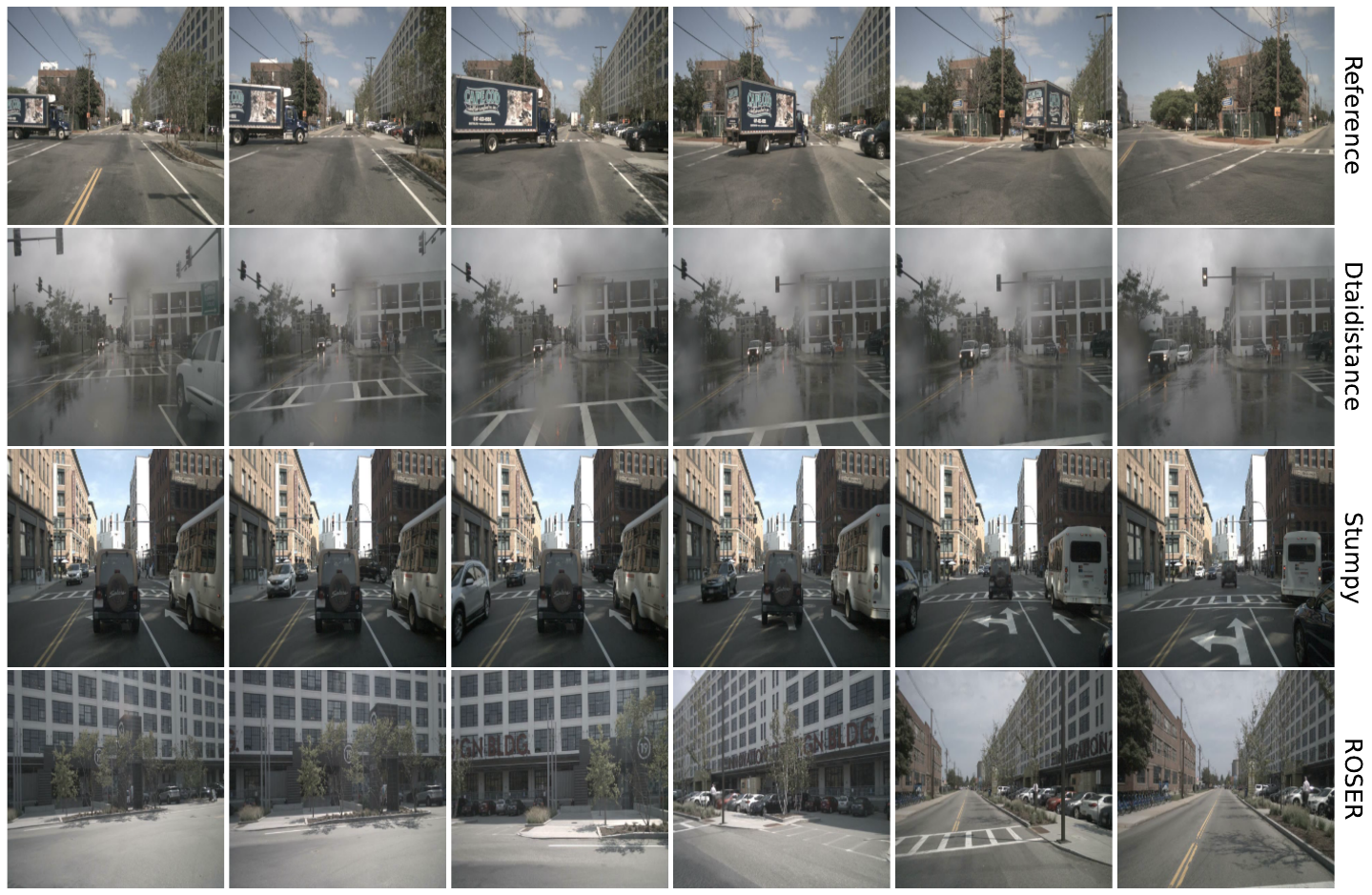}
    \caption{nuScenes qualitative results for left turn maneuver. ROSER successfully retrieves similar task while Dtaidistance retrieves a regular stop and Stumpy retrieves a straight driving scene.}
    \label{fig:quality-nuscene-left-turn}
\end{figure}
\clearpage

% --- Right Turn ---
\subsubsection{Right Turn}
\begin{figure}[!ht]
    \centering
    \includegraphics[width=\linewidth]{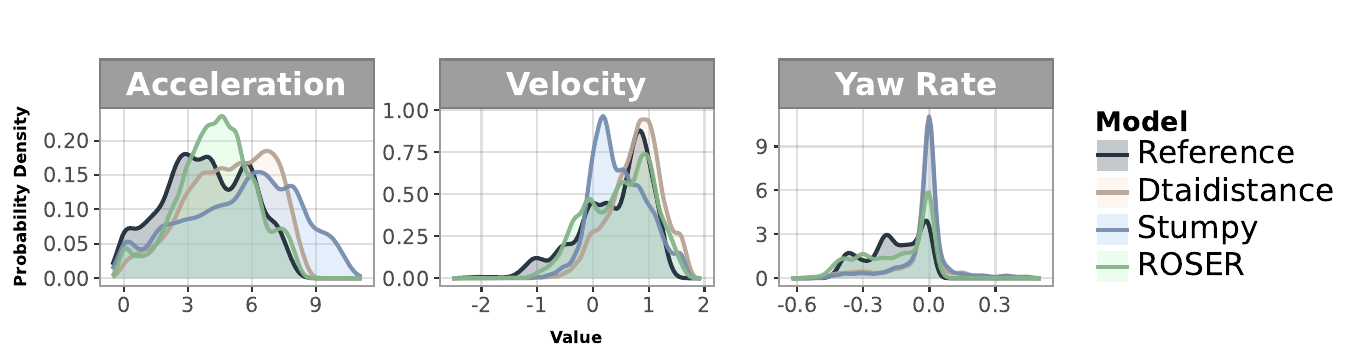}
    \caption{Feature-level distribution visualization for right turn maneuver in the nuScenes benchmark.}
    \label{fig:dist-nuscene-right_turn}
\end{figure}

\begin{figure}[!ht]
    \centering
    \includegraphics[width=\linewidth]{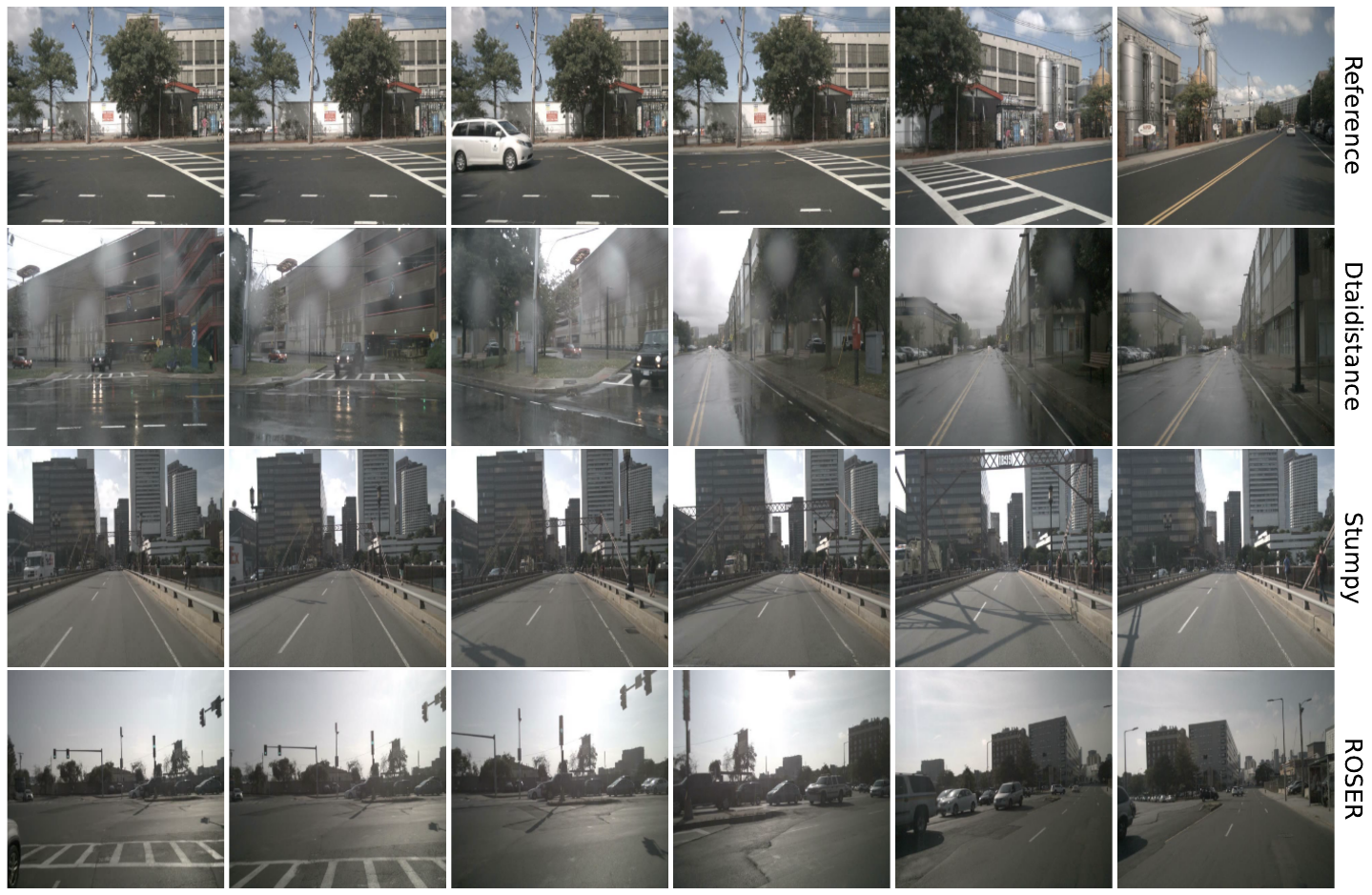}
    \caption{nuScenes qualitative results for right turn maneuver. ROSER and Dtaidistance successfully retrieves similar task while Stumpy retrieves a straight driving scene.}
    \label{fig:quality-nuscene-right-turn}
\end{figure}
\clearpage

% --- Straight Driving ---
\subsubsection{Straight Driving}
\begin{figure}[!ht]
    \centering
    \includegraphics[width=\linewidth]{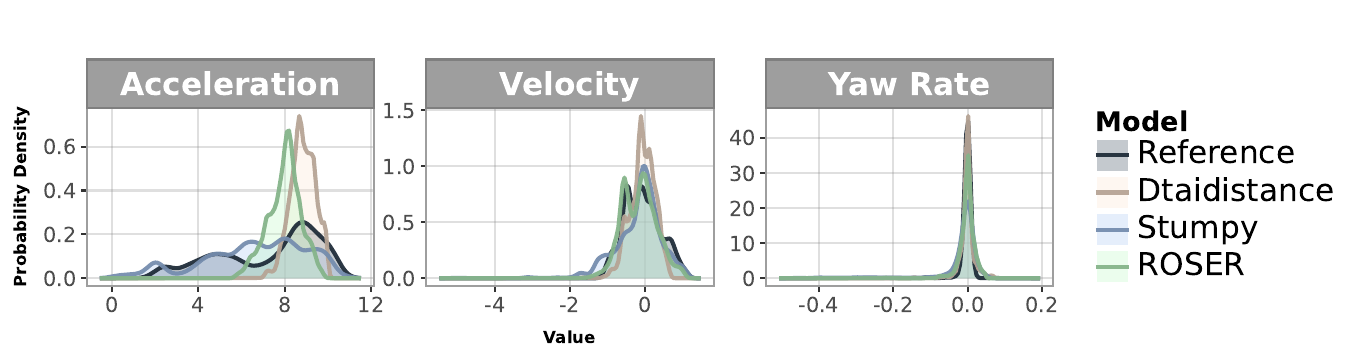}
    \caption{Feature-level distribution visualization for straight driving maneuver in the nuScenes benchmark.}
    \label{fig:dist-nuscene-straight_driving}
\end{figure}

\begin{figure}[!ht]
    \centering
    \includegraphics[width=\linewidth]{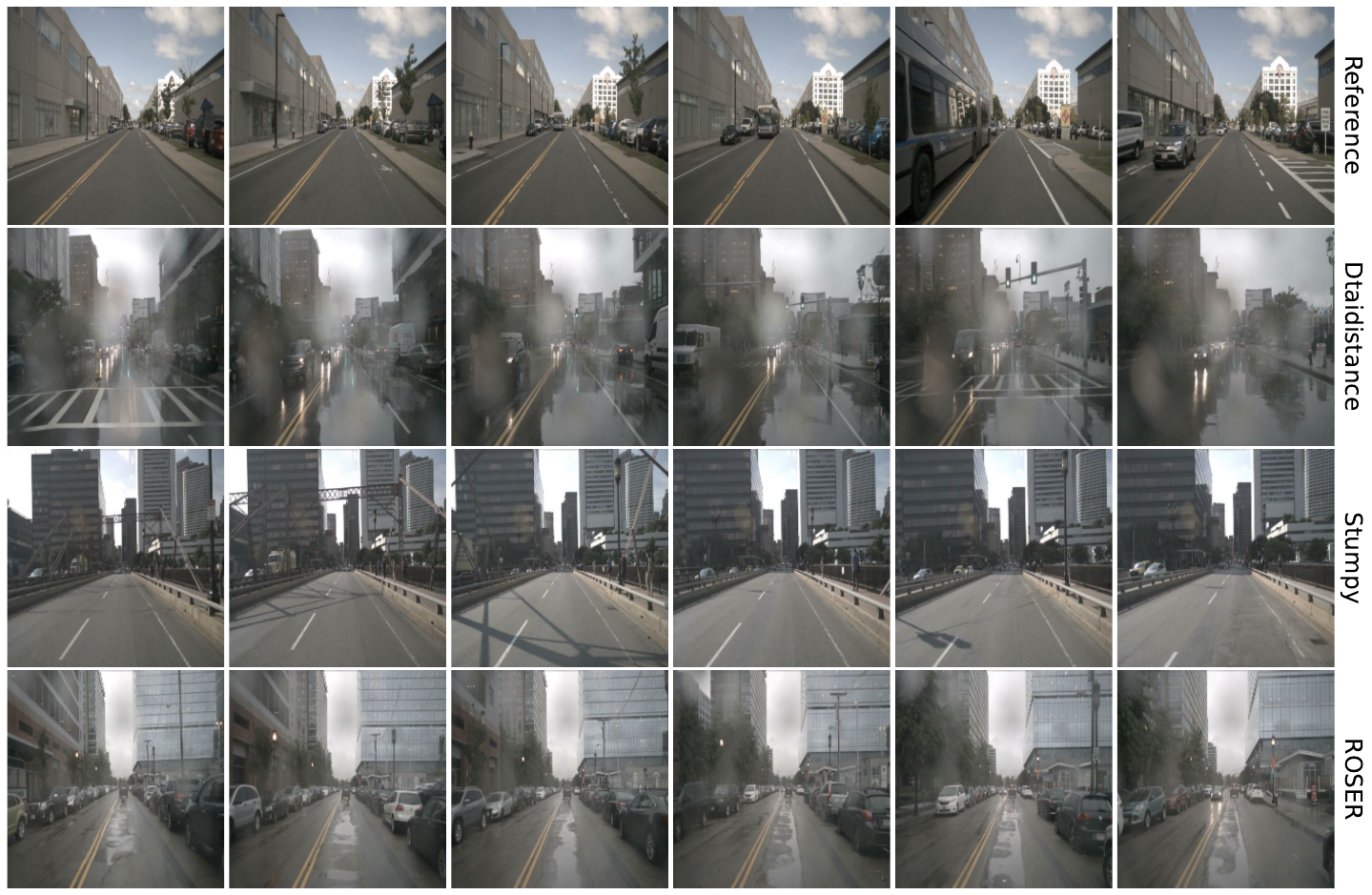}
    \caption{nuScenes qualitative results for straight driving maneuver. All models successfully retrieve similar scene for this task.}
    \label{fig:quality-nuscene-straight-driving}
\end{figure}
\clearpage

% --- Regular Stop ---
\subsubsection{Regular Stop}
\begin{figure}[!ht]
    \centering
    \includegraphics[width=\linewidth]{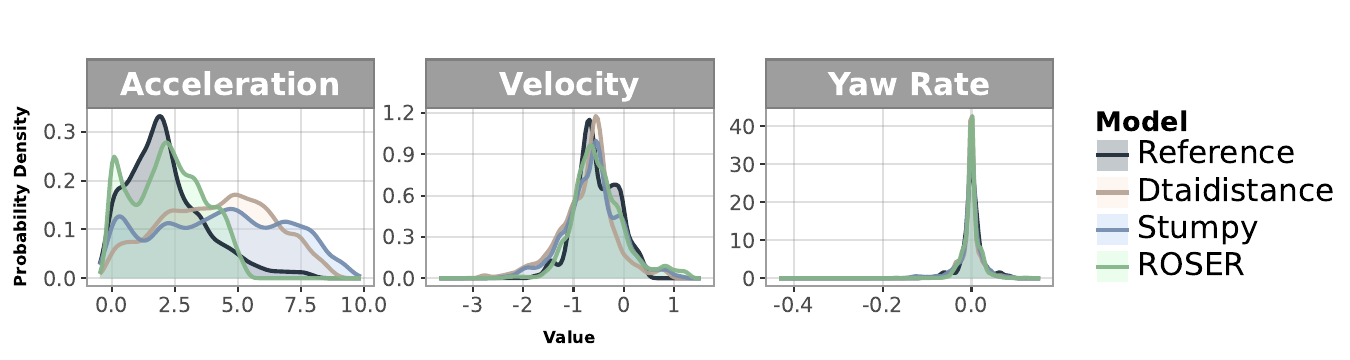}
    \caption{Feature-level distribution visualization for regular stop maneuver in the nuScenes benchmark.}
    \label{fig:dist-nuscene-regular_stop}
\end{figure}

\begin{figure}[!ht]
    \centering
    \includegraphics[width=\linewidth]{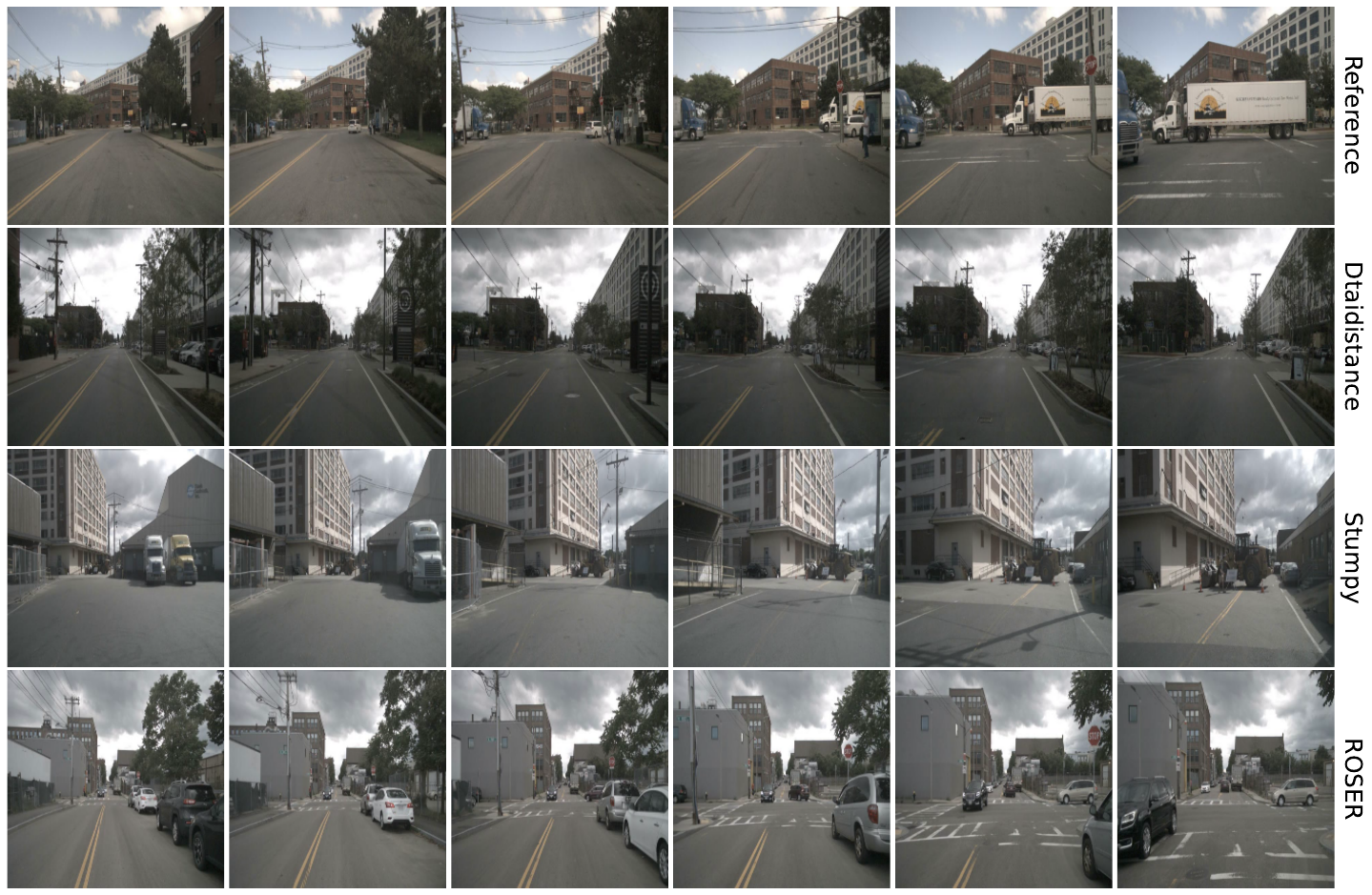}
    \caption{nuScenes qualitative results for regular stop maneuver. ROSER successfully retrieves similar task as the vehicle stops at the crosswalk while Dtaidistance and Stumpy retrieve straight driving scenes.}
    \label{fig:quality-nuscene-regular-stop}
\end{figure}
\clearpage

\end{document}